\documentclass[letterpaper]{article} % DO NOT CHANGE THIS
\usepackage{aaai25}  % DO NOT CHANGE THIS
\usepackage{times}  % DO NOT CHANGE THIS
\usepackage{helvet}  % DO NOT CHANGE THIS
\usepackage{courier}  % DO NOT CHANGE THIS
\usepackage[hyphens]{url}  % DO NOT CHANGE THIS
\usepackage{graphicx} % DO NOT CHANGE THIS
\usepackage{natbib}  % DO NOT CHANGE THIS AND DO NOT ADD ANY OPTIONS TO IT
\usepackage{caption} % DO NOT CHANGE THIS AND DO NOT ADD ANY OPTIONS TO IT
\usepackage{nicefrac}       % compact symbols for 1/2, etc.
\usepackage{microtype}      % microtypography
\usepackage{algorithm}
\usepackage{algorithmicx}
\usepackage{amsmath}
\usepackage{algpseudocode}

\usepackage{booktabs}
\usepackage{adjustbox}
\usepackage{amsfonts}
\usepackage{xcolor}

\newtheorem{assumption}{Assumption}

\newtheorem{theorem}{Theorem}

\newcommand{\bE}{\mathbb{E}}

\usepackage{newfloat}
\usepackage{listings}
\DeclareCaptionStyle{ruled}{labelfont=normalfont,labelsep=colon,strut=off} % DO NOT CHANGE THIS
\floatstyle{ruled}
\newfloat{listing}{tb}{lst}{}
\floatname{listing}{Listing}
\title{Constraint-Adaptive Policy Switching for Offline Safe Reinforcement Learning}

\author{Yassine Chemingui\textsuperscript{\rm 1}, Aryan Deshwal\textsuperscript{\rm 2}, Honghao Wei\textsuperscript{\rm 1}, Alan Fern\textsuperscript{\rm 3}, Janardhan Rao Doppa\textsuperscript{\rm 1}\\ % All authors must be in the same font size and format. Use \Large and \textbf to achieve this result when breaking a line
\textsuperscript{\rm 1} {\normalfont \normalsize Washington State University} \textsuperscript{\rm 2} {\normalfont \normalsize University of Minnesota} \textsuperscript{\rm 3} {\normalfont \normalsize Oregon State University}\\ 
{\normalfont \normalsize \texttt{ \{yassine.chemingui,honghao.wei,jana.doppa\}@wsu.edu, adeshwal@umn.edu, alan.fern@oregonstate.edu}}
}

\begin{document}

\maketitle

% New (Old + Yassine)
\begin{abstract}
Offline safe reinforcement learning (OSRL) involves learning a decision-making policy to maximize rewards from a fixed batch of training data to satisfy {\em pre-defined} safety constraints. However, adapting to varying safety constraints during deployment without retraining remains an under-explored challenge. 
To address this challenge, we introduce constraint-adaptive policy switching (CAPS), a wrapper framework around existing offline RL algorithms. 
During training, CAPS uses offline data to learn multiple policies with a shared representation that optimize different reward and cost trade-offs. During testing, CAPS switches between those policies by selecting at each state the policy that maximizes future rewards among those that satisfy the current cost constraint. Our experiments on 38 tasks from the DSRL benchmark demonstrate that CAPS consistently outperforms existing methods, establishing a strong wrapper-based baseline for OSRL.  The code is publicly available at \url{https://github.com/yassineCh/CAPS}.
\end{abstract}

\section{Introduction}
Online reinforcement learning (RL) has shown great successes in applications where the agent can continuously interact with the environment to collect new feedback data to improve its decision-making policy \cite{mnih2015human, silver2018general, li2017deep}. However, in many real-world domains such as agriculture, smart grid and healthcare, executing exploratory decisions is costly and/or dangerous, but we have access to pre-collected datasets. Offline RL \cite{levine2020offline} is an emerging paradigm to learn decision policies {\em solely} from such offline datasets, eliminating the need for additional interaction with the environment. The key challenge in offline RL is addressing the distributional shift between the state-action distribution of the offline dataset and the learned policy, as this shift results in extrapolation error. Existing offline RL methods typically employ the principle of pessimism in the face of uncertainty (i.e., some form of penalization to deal with out-of-distribution states/actions) to address this challenge \cite{levine2020offline}.

In many safety-critical domains (e.g., agriculture and smart grid), the agent behavior needs to also satisfy some cost constraints in addition to maximizing reward. Most of the prior work on offline RL focuses on the unconstrained setting where the policy is optimized to maximize rewards. 
There is relatively less work on offline constrained/safe RL and the majority of these methods assume that the cost constraint is known during the training phase (as elaborated in the related work section). However, in many applications, the cost constraints can change at deployment depending on the use-case scenario. 
For example, in agricultural drones, the reward is the area covered during spraying, and the cost is the battery/power consumption. The goal is to maximize efficiency for a given agricultural field by optimizing coverage based on the available battery power.

This paper studies the problem of offline safe RL (OSRL) where the safety/cost constraints can vary after deployment. We propose {\em Constraint-Adaptive Policy Switching (CAPS)}, an algorithm that can be wrapped around existing offline RL methods and is easy to implement. CAPS handles the challenge of unknown cost constraint threshold as follows. At training time, we learn multiple policies with a shared representation with the goal of achieving different reward and cost trade-offs: one centered on reward maximization, one on cost minimization, and other policies optimized for varied linear scalarizations of costs and rewards. At testing time, CAPS switches between the decisions from the trained policies at each state, selecting the one that maximizes future rewards from the ones that satisfy the cost constraint. We also theoretically analyze the conditions under which CAPS decision-making process has safety guarantees. 

We perform experiments on 38 tasks from the DSRL benchmark \cite{liu2023datasets} to compare CAPS with state-of-the-art OSRL baselines and our main findings are as follows. First, CAPS wrapped around two qualitatively different offline RL algorithms exhibits safe behavior (first-order objective) on a larger fraction of tasks compared to baselines across different cost threshold constraints. Prior methods struggle even with known cost constraints, potentially because of estimation errors in value functions and/or instability in the Lagrange multiplier. Second, the implicit Q-learning instantiation of CAPS demonstrates safe performance in \textbf{34/38 tasks (89\%)} and achieves the highest rewards in \textbf{18} of those tasks compared to the baselines.. Third, CAPS using shared representation performs significantly better than CAPS with independently trained policies. Finally, increasing the number of policies improves the overall performance of CAPS, but CAPS with two policies (one for reward maximization and one for cost minimization) is competitive, has the advantage of removing extra hyperparameters, and stands as a strong minimalist approach for effective offline safe RL.

\vspace{1ex}

\noindent {\bf Contributions.} The key contribution is the development and evaluation of CAPS framework for offline safe RL with varying cost constraints. Specific contributions include:

\begin{itemize}
    \item Development of CAPS to handle any test-time cost threshold constraint. CAPS can be wrapped around existing offline RL algorithms and easy to implement.
    \item Theoretical analysis to understand the conditions under which CAPS decision policy has safety guarantees. 
    \item Experiments and ablations on 38 DSRL tasks, demonstrate the effectiveness of CAPS over prior methods.
\end{itemize}

\section{Problem Setup}

RL problems with safety constraints are naturally formulated within the Constrained Markov Decision Process (CMDP) framework. A CMDP is defined by a tuple \(M = (S, A, P, r, c, \mu_0)\), where \(S\) is the state space, \(A\) is the action space, \(P: S \times A \times S \to [0, 1]\) is the state transition probability function, \(r: S \times A \to \mathbb{R}\) defines the reward function, and \(c: S \to [0, C_{\text{max}}]\) quantifies the costs associated with states, where \(C_{\text{max}}\) is the maximum possible cost, and $\mu_0: S \rightarrow [0, 1]$ is the initial state distribution. For simplicity of notation, we define cost function on states noting that it is easy to define an equivalent CMDP with costs associated with state-action pairs. 

In offline safe RL problems, we are given a fixed pre-collected dataset $\mathcal{D}$=$\{(s, a, r, c, s')_i\}_{i=1}^{n}$ from one or more (unknown) behavior policies, where each training example $i$ contains the action $a$ taken at state $s$, reward received $r$, cost incurred $c$, and the next state $s'$. The goal is to learn a policy \(\pi: S \to A\) from the offline dataset \(\mathcal{D}\) to maximize the expected reward while satisfying a specified cost/safety constraint. This problem is mathematically formulated as:
\begin{align*}
\max_{\pi} \mathbb{E}_{\tau \sim \pi}[R(\tau)] \quad \mathrm{subject\ to} \quad \mathbb{E}_{\tau \sim \pi}[C(\tau)] \leq \kappa,
\end{align*}
Here, \(\kappa \in [0, +\infty)\) is the cost threshold for safety constraint, \(\tau = \{s_1, a_1, r_1, c_1, \ldots, s_T, a_T, r_T, c_T\}\) denotes a trajectory sampled by executing the policy $\pi$, \(T\) is the length of episode, \(R(\tau) = \sum_{t=1}^T r_t\) is the total accumulated reward, and \(C(\tau) = \sum_{t=1}^T c_t\) is the total incurred cost.

The majority of prior work on OSRL assumes that the cost threshold $\kappa$ for safety constraint is known at the training time and remains same at deployment. However, this assumption is violated in many real-world applications where the safety requirements can change during deployment based on the operating conditions or environmental factors. For example, in agricultural crop management, the goal may be to maximize yield under constraints on water and/or nitrogen application depending on time-varying regulations. Additionally, in a decision-support scenario, it can be useful for decision makers to quickly explore the impact of different cost thresholds without requiring retraining.

Other application domains where this problem setting arises include healthcare, power systems, and robotics. Consequently, 
OSRL methods that learn a single policy tailored to a specific cost constraint are ill-equipped to handle varying test-time cost constraints. 

The goal of this paper is to fill this knowledge gap in the current literature by studying a variant of the OSRL problem where the cost constraint $\kappa$ can vary during deployment. 

\section{Related Work}
\vspace{1ex}
\noindent \textbf{Online Safe RL.} RL with safety constraints in the online setting, where the agent can interact with the environment to collect feedback data, is extensively studied \cite{garcia2015comprehensive, gu2022review}. However, most of these methods assume a fixed constraint threshold during both training and deployment. 
Constraint-conditioned policy optimization \cite{yao2024constraint}, addresses this limitation and adapts to changing constraints 
using two strategies: approximating value functions under unseen constraints and encoding arbitrary constraint thresholds during policy optimization.

\vspace{1ex}

\noindent \textbf{Offline RL.} The goal in offline RL is to learn effective policies from fixed static datasets without any additional interaction with the environment. Prior work has tackled the key challenge of distribution shift in offline RL \cite{levine2020offline, offline_rl_survey_2}. Some representative techniques include incorporating regularization in value function estimation \cite{fujimoto2021minimalist, kostrikov2022offline, kumar2019stabilizing, lyu2022mildly, yang2022rorl}, sequential modeling formulations \cite{janner2021offline, wang2022bootstrapped}, uncertainty-aware methods \cite{an2021uncertainty, bai2022pessimistic}, divergence based policy constraints \cite{wu2019behavior, jaques2019way, wu2022supported}, and extensions of model based RL \cite{kidambi2020morel, yu2020mopo, rigter2022rambo} and imitation learning \cite{xu2022policy}. SfBC \cite{chen2023offline} and IDQL \cite{hansen2023idql} filter candidate actions using Q-function evaluations and reweight them for selection. In contrast, CAPS avoids reweighting by directly selecting the best safe action using two Q-functions. PEX \cite{PEX} is a offline-to-online RL approach that adds a new expansion policy to the offline policy, unlike CAPS, which dynamically switches between pre-trained policies for safety.

\vspace{1ex}

\noindent \textbf{Offline Safe RL.} Recent work has studied the offline safe RL problem, aiming to learn a safe policy from a given offline dataset \cite{liu2023datasets}. These methods include finding cost-conservative policies for better actual constraint satisfaction via some form of constrained policy optimization formulation \cite{polosky2022constrained, lee2022coptidice}, Lagrangian based approaches to handle cost constraints \cite{xu2022constraints}, and the constrained decision transformer (CDT) \cite{liu2023constrained}, which exploits the advances in sequential modeling by learning from a dataset of trajectories. TREBI \cite{lin2023safe} and FISOR \cite{zheng2024safe} leverage diffusion models for creating safe policies: TREBI  generates safe trajectories, while FISOR uses a diffusion actor to select actions within feasible regions.
However, these methods, with the exception of CDT, rely on fixed cost thresholds during training, limiting their flexibility for deployment scenarios with varying cost threshold constraints.

\section{Constraint-Adaptive Policy Switching}

In this section, we first describe the reasoning procedure behind the  {\em Constraint-Adaptive Policy Switching (CAPS)} approach of handling unknown constraints at testing time. Next, we describe the training process and the design choices that enable effective decision-making in CAPS.

\subsection{Decision-Making in CAPS}

One way to handle dynamically changing cost constraints would be to: 1) train a large set of $K$ policies that span a wide spectrum of cost-reward trade-offs, and 2) at each test-time state, select the best of those policies that satisfies the current cost constraint. While this makes sense in concept, there are at least two key practical issues. First, there is a high computational cost involved in optimizing $K$ policies via safe offline RL. Second, and more importantly, step (2) requires accurately assessing the safety of a given policy and also selecting the best among the safe ones. This requires accurate cost and value functions for each policy, which would need to be estimated using off-policy evaluation (OPE) techniques with the available dataset. Unfortunately, as our experimental results show, current OPE techniques are not reliable enough to ensure test-time safety \cite{offline_rl_survey_2}. The key idea of CAPS is to address these two issues by creating a set of $K$ diverse policies at a much smaller computational cost, while avoiding a reliance on OPE to provide test-time safety.

CAPS relies on two key components for its decision-making to handle unknown cost constraints. First, a set of $K \geq 2$ 
policies $\mathcal{P} = \{\pi^r,\pi^1, \dots, \pi^{K-2}, \pi^c\}$ that are trained using the offline dataset $\mathcal{D}$ to cover actions with  different reward and cost trade-offs. 
At one extreme, $\pi^r$ is trained to maximize reward via standard Offline-RL, while ignoring the cost objective. 
At the other extreme, $\pi^c$ is trained to minimize cost via standard Offline-RL, while ignoring reward. In this sense, $\pi^c$ is maximally safe, while $\pi^r$ is cost agnostic. 
The remaining $K-2$ policies are trained with the intent of covering actions that achieve a variety of reward and cost trade-offs, while avoiding the full computational cost of Offline-RL for each one. 
Second, CAPS retains from Offline-RL the learned Q-functions $Q_t^r$ and $Q_t^c$ 
which we assume are non-stationary given the finite time horizon: 

\begin{align}
    Q_t^r(s,a) = & \bE[\sum_{t}^T r_t \vert s_t=s,a_t=a] \\
    Q_t^c(s,a) = & \bE[\sum_{t}^T c_t \vert s_t=s,a_t=a],
\end{align}
where the expectation is taken over the corresponding policy with respect to randomness from reward/cost functions and state transitions. Intuitively, CAPS will use $Q_t^c$ to help filter out unsafe decisions from the $K$ policies and $Q_t^r$ is employed to select the best decision from the unfiltered ones. Note that these Q-functions are an artifact of policy optimization, rather than OPE of an arbitrary policy. This is a subtle, but important, aspect of CAPS, since it has been observed that value estimates of arbitrary policies via OPE tends to be less reliable \cite{offline_rl_survey_2}. 

\vspace{1ex}

Given $K$ policies $\mathcal{P}$=$\{\pi^r, \pi^1, \dots, \pi^{K-2},\pi^c\}$, two Q-functions $Q^r$ and $Q^c$, cost threshold constraint $\kappa$, and accumulated cost $c_{\leq t}$ up to current state $s_t$, CAPS selects an action as follows:

\begin{enumerate}
    \item \textbf{Filter}: Identify the subset of feasible actions $\mathcal{F}_t \subseteq \{\pi^r(s_t), \pi^1(s_t), \dots, \pi^{K-2}(s_t),\pi^c(s_t)\}$ selected by the $K$ policies and are safe in $s_t$ according to $Q^c$ and cost threshold $\kappa$:
     
    \begin{align}
        \mathcal{F}_t = \{\pi(s_t) \, \mid \, Q_t^c (s_t, \pi(s_t)) + c_{\leq t} \leq \kappa , \pi \in \mathcal{P} \}
    \end{align}
    \item \textbf{Select}: Among the feasible actions $\mathcal{F}_t$, pick the one that that maximizes the estimated reward according to $Q^r$: %, i.e.,
    \begin{align}
        a^*_t = \arg \max_{a \in \mathcal{F}_t} Q_t^r (s_t, a)
    \end{align}
\end{enumerate}
If the filter step results in an empty set of feasible actions $(\mathcal{F}_t = \emptyset)$, CAPS selects the action associated with the policy that minimizes the estimated cost: \begin{align*}
    a^*_t &= \pi^*_t (s_t), \quad
    \text{where} \quad \pi^*_t &= \arg \min_{\pi \in \mathcal{P}} Q_t^c (s_t, \pi(s_t))
\end{align*}

Intuitively, the first filtering step keeps any action selected by a policy such that it is safe to take that action and then follow $\pi^c$ thereafter. That is, after the action is taken there is an assurance in expectation that $\pi^c$ is a safe fallback policy. The selection step then heuristically selects the action that looks best from a reward maximization point of view. 

\subsection{Training Algorithm for CAPS}

Given an offline dataset $\mathcal{D} = \{(s_i, a_i, r_i, c_i, s^{\prime}_i)\}_{i=1}^n$, the CAPS approach aims to efficiently learn a set of policies $\mathcal{P}$=$\{\pi^r, \pi^1, \dots, \pi^{K-2},\pi^c\}$ and the two Q-value functions $Q^r$ and $Q^c$ that yield strong test time performance. A key feature of CAPS is that training is done via a reduction to standard Offline-RL. In particular, CAPS can be combined with any Offline-RL algorithm that produces a Q-function and a mechanism to extract a policy for that Q-function. In what follows, we first describe the high-level training schema and then provide two concrete instantiations that are used in our experiments. Finally, we discuss the design choice of network architecture, including shared structure between the $K$ policies.

\subsubsection{Reduction to Offline RL.}
The training approach for CAPS can be wrapped around existing offline RL algorithms with minimal changes, making it straightforward to implement. 
The general recipe consists of the following two steps:
    \begin{enumerate}
        \item Train the reward-only value funcion \(Q^r\) and cost-only value function \(Q^c\). For \(Q^c\), we train using cost data from the offline dataset by defining the rewards as costs.
        \item Extract the reward maximizing policy \(\pi^r\) from \(Q^r\), the cost minimizing policy \(\pi^c\) from \(Q^c\), and the other $K-2$, \(\pi^k\), policies with different reward and cost trade-offs from  the mixture Q-value function $Q^r -  \lambda_k Q^c$ following the same policy extraction procedure in the offline RL method, where $\lambda_k$ is the scalarization parameter.
    \end{enumerate}
Thus, the approach requires just two full runs of an Offline RL algorithm and $K$ runs of the much cheaper policy extraction step. It is important to note that the Q-functions for costs and rewards are not learned for each $\pi^k$, which means that the precise cost-reward trade-off of each policy is unknown. Thus, this approach to producing each $\pi^k$ is a fast heuristics for identifying a set of policies that are likely to span a range of cost-reward trade-offs. Another advantage of this reduction approach is that we can leverage better offline RL algorithms to further improve the performance of CAPS and our experiments demonstrate this advantage. While it is possible to apply OPE methods to learn these Q-functions and use them for decision making, our experiments show that while safety is maintained, the rewards are significantly lower compared to our CAPS decision procedure.

\subsubsection{Two CAPS Instantiations. \label{sec:instantiations}}
Our two instantiations of CAPS training are based on two state-of-the-art offline RL methods: IQL and SAC+BC.

\vspace{1ex}

\textit{IQL Instantiation:} We employ the Implicit Q-Learning (IQL) algorithm \cite{kostrikov2022offline} as follows. Given the offline dataset \( \mathcal{D} = \{(s_i, a_i, r_i, c_i, s'_i)\}_{i=1}^n \), we additionally train a value function \( V^c \) and a Q-value function \( Q^c \) for costs. These functions are trained similarly to the original IQL reward value \( V^r \) and Q-value \( Q^r \) functions by minimizing the expectile loss, thereby avoiding explicit value queries for unseen actions.
Next, we extract each policy $\pi \in \mathcal{P}$ by maximizing the following objective.
\begin{equation}
J_{\pi^k}(\phi) = \mathbb{E}_{s, a \sim \mathcal{D}} \left[ \exp(\beta (A^r - \lambda_k A^c)) \log\pi_\phi(a|s)
\right]
\end{equation}

\noindent where the advantages are defined as:
\begin{equation}
A^r = Q_\theta^r(s, a) - V_\psi^r(s), \quad
A^c = Q_\theta^c(s, a) - V_\psi^c(s)
\tag{9}
\end{equation}

where $\theta$ and $\psi$ correspond to the parameters of the Q-value and value function network representation. Note that for $\pi^r$ and $\pi^c$, we only use $A^r$ and $-A^c$ respectively.

\vspace{1ex}

\textit{SAC+BC Instantiation:} \cite{fujimoto2021minimalist}) recently described a general framework for developing offline RL algorithms by combining an off-policy RL with a behavior cloning (BC) regularization term. We follow the same principle but employ soft-actor-critic (SAC) \cite{haarnoja2018soft} algorithm as the off-policy RL component instead of the TD3 algorithm used in the original paper. Following the offline reduction recipe, we update the actor network to incorporate $K$ heads and define an additional Q-value function for costs. We then learn each policy $\pi^k$ by maximizing the following objective:
\begin{equation}
\begin{aligned}
J_{\pi^k}(\phi) = \mathbb{E}_{s, a \sim \mathcal{D}, \tilde{a} \sim \pi_{\phi}(\cdot \vert s)} \Big[ &  Q^k(s, \tilde{a})
 - \alpha \log \pi_{\phi}(\tilde{a} \vert s) \\
 &- (\tilde{a} - a)^2 \Big]
\end{aligned}   
\end{equation}

\noindent where $Q^k(s, \tilde{a}) = Q^r_{\theta^r}(s, \tilde{a}) - \lambda_k Q^c_{\theta^c}(s, \tilde{a})$. For $\pi^r$ and $\pi^c$, we use $Q^r$ and $-Q^c$ to update their corresponding heads.

\subsubsection{Improving CAPS via Shared Actor Representation.}
To enhance knowledge transfer and/or training efficiency, we utilize a shared neural network architecture for the $K$ different policies. Instead of training separate networks, we employ a common body with parameters $\phi_s$ that learns a unified state representation $f_{\phi_s}(s)$. This body is paired with distinct output heads (one for each of the $K$ policies), each with its own parameters $\phi_k$, which specialize in predicting the action distributions for their respective policies: $\pi^k(a|s) = g_{\phi_k}(f_{\phi_s}(s)), \forall k \in [K]$. 

There are two synergistic benefits of a shared policy representation. First, this approach captures general features relevant to both reward and cost objectives, leading to knowledge transfer across policies. 
Second, CAPS requires only one round of training to adapt to different cost constraint thresholds. This saves significant training time compared to methods that need re-training for each cost constraint.

\section{Safety Guarantee}

This section describes conditions under which the CAPS decision-making process has safety guarantees. For simplicity, we consider the case where the optimal-cost Q-function is perfectly estimated, noting that it is straightforward to modify our result to account for bounded estimation error. 

We focus on the finite-horizon setting with horizon $T$ and let $Q^c_t$ and $V^c_t$ denote the cost optimal Q and value functions for time step $t \in \{0,\ldots, T\}$ respectively. Here we treat these functions as positive cost accumulation functions that are minimized during optimization. We consider policies of the form $\pi_t(s,c_{<t})$ where $s$ is the current state and $c_{<t}$ is the accumulated cost before arriving at time step $t$. For a given cost safety bound $\kappa$, our guarantee applies to any policy that is $\kappa$-admissible.
\begin{assumption}
\label{assumption:admissible}
A policy $\pi$ is $\kappa$-admissible if for any state $s$, any time step $t$, and any accumulated cost $c_{<t}$, we have $Q^c_t(s,\pi_t(s,c_{<t})) \leq \max\{V^c_t(s), \kappa-c_{<t}\}$.
\end{assumption}
Note that by assuming a perfect $Q^c$, the switching policy defined for CAPS is $\kappa$-admissible since it either picks an action $a$ such that $Q_t^c(s,a) \leq \kappa-c_{<t}$ if possible, or resorts to choosing the least cost action, which has a value of $V^c_t(s)$ since $Q^c$ and $V^c$ are cost optimal. This is true of CAPS for any set of $K$ policies $\mathcal{P}$ that contains the cost optimal policy $\pi_c$ and any reward value function $Q^r_t$.  

Our safety guarantees are specified in terms of a parameter $\epsilon$ that characterizes the underlying Markov decision process (MDP). In particular, safety in terms of expected cost is challenging when stochastic state transitions can result in a set of possible next states that have wildly different values of $V^c$. Indeed, counter examples to safety can be constructed that involve transitions with very small probabilities of reaching enormous values of $V^c$ while the remaining probability mass results in very small costs. For this purpose, we define the notion of \emph{optimal-cost variation}. 

\begin{assumption}
\label{assumption:variation}
Given an MDP $M$, let $N(s,a) = \{s' \;|\; P(s,a,s') > 0\}$ be the set of possible next states after taking action $a$ in state $s$. $M$ has an optimal-cost variation of $\epsilon$ if each state-action-timestep tuple $(s,a,t)$ satisfies $\max_{s'\in N(s,a)} V^c_t(s') - \min_{s'\in N(s,a)} V^c_t(s') \leq \epsilon$.     
\end{assumption}
Note that any deterministic MDP has $\epsilon = 0$. Also note that this assumption places no constraints between the optimal costs of states resulting from different $(s,a,t)$ tuples. We can now state our main result which bounds the total cost of a policy $\pi$, denoted $V^{\pi,c}_t$. 
\begin{theorem}
\label{thm:bound}
    For any MDP $M$ with optimal-cost variation $\epsilon$ and any $\kappa$-admissible policy $\pi$, we have that for any state $s$, time step $t$, and accumulated cost $c_{<t}$, $$V^{\pi,c}_t(s)\leq \max\{V^c_t(s),\kappa-c_{<t}\} + (T-t)\epsilon$$
\end{theorem}
See the supplementary material for the full proof. 

This result says that up to a constant factor either the expected total cost of policy $\pi$ will be bounded by $\kappa$ over a horizon of $T$ or if that is not possible, it will be bounded by the lowest possible expected cost $V^c_t(s)$. Based on this result, if one has an estimate of $\epsilon$ for the system, an appropriate value of $\kappa$ can be selected to provide a desired guarantee. 
The constant factor $(T-t)\epsilon$ is simply a constant of $\epsilon$ accumulated over the remaining horizon $T-t$. Note that for time step 0, we get a bound of $\kappa$ when it is possible to achieve that safety factor. Also, note that in the deterministic case, the bound is tight since $\epsilon = 0$.

\section{Experiments and Results}

This section presents experimental evaluation of the CAPS approach and comparison with state-of-the-art methods. 

\subsection{Experimental Setup}

\vspace{1ex}

\noindent {\bf Benchmarks.} We employ 38 sequential decision-making benchmarks of varying difficulty from Safety-Gymnasium \cite{ray2019benchmarking, ji2023omnisafe}, Bullet-Safety-Gym \cite{gronauer2022bullet}, and MetaDrive \cite{li2022metadrive} within the DSRL framework \cite{liu2023datasets}. Further details are provided in Appendix \ref{ap:envs_desc}.

\vspace{1ex}

\noindent {\bf Configuration of CAPS.} We consider two instantiations of CAPS training with different offline RL methods: \texttt{CAPS(IQL)} employs IQL \cite{kostrikov2022offline} and \texttt{CAPS(SAC+BC)} employs soft-actor-critic based off-policy RL combined with a behavior cloning regularization term \cite{haarnoja2018soft, fujimoto2021minimalist} as described in the technical section. We consider CAPS with different number of policies $K \in \{ 2, 4, 8\}$ to determine the best configuration in terms of computational cost and quality of decisions. We provide the details of the neural network structure used for value and Q-functions, policy heads, and hyper-parameters in the Appendix \ref{ap:caps_hyper}.

\vspace{1ex}

\noindent {\bf Baseline methods.} We compare CAPS with several state-of-the-art baseline methods. { {\bf 1)} \texttt{BC} (Behavior Cloning)} serves as a straightforward baseline where policies are learned by mimicking the behavior observed in the training data. {\bf 2)} \texttt{BEAR-Lag} integrates BEAR \cite{wu2019behavior}, an offline RL method, with a Lagrangian approach to address safety constraints. {{\bf 3)} Constraints-penalized Q-learning (\texttt{CPQ})} treats out-of-distribution actions as unsafe and updates the Q-value function using only safe state-action pairs \cite{xu2022constraints}. { {\bf 4)} \texttt{COptiDICE}} builds upon the distribution correction estimation (DICE) method to incorporate cost constraints in offline RL \cite{lee2022coptidice} . {{\bf 5)} \texttt{CDT}} employs a decision transformer framework to consider safety constraints for safe decision-making \cite{liu2023constrained}. This is the only baseline that allows varying cost constraints at test-time. We run each over three seeds.

\vspace{1ex}

\noindent {\bf Evaluation methodology.} We evaluate CAPS and the above-mentioned baseline methods using normalized return and normalized cost, where a normalized cost below 1 indicates that safety constraints are satisfied. In line with the DSRL guidelines, safety is our primary evaluation metric and our goal is to maximize rewards while adhering to cost constraints. We evaluate each algorithm using three different target cost threshold configurations, three random seeds, and twenty episodes. The configurations are: \{10, 20, 40\}, \{20, 40, 80\}, where the second cost limit set applies to the more challenging Safety-Gymnasium environments. We compute the average normalized reward and cost to assess the performance. Note that, for all baseline methods except CDT, a separate agent is trained for each cost threshold.

\subsection{Results and Discussion}

\begin{table*}[ht!]
\centering
\caption{Results for normalized rewards and costs. The cost threshold is 1. The $\uparrow$ symbol denotes that higher the rewards, the better. The $\downarrow$ symbol denotes that the lower the costs (up to threshold 1), the better. Each value is averaged over 3 distinct cost thresholds, 20 evaluation episodes, and 3 random seeds. \textbf{Bold}: Safe agents whose normalized cost $\leq$ 1. \textcolor{gray}{Gray}: Unsafe agents. \textbf{\textcolor{blue}{Blue}}: Safe agent with the highest reward. We compare CAPS instantiations with other baselines, not with each other.}
\label{tab:main_results}
\begin{adjustbox}{max width=\textwidth}
\begin{tabular}{lcccccccccccccc}
\toprule
Task & \multicolumn{2}{c}{BC} & \multicolumn{2}{c}{BEAR-Lag} & \multicolumn{2}{c}{CPQ} & \multicolumn{2}{c}{CDT} & \multicolumn{2}{c}{COptiDICE} & \multicolumn{2}{c}{CAPS (SAC+BC)} & \multicolumn{2}{c}{CAPS (IQL)} \\
\cmidrule(lr){2-3} \cmidrule(lr){4-5} \cmidrule(lr){6-7} \cmidrule(lr){8-9} \cmidrule(lr){10-11} \cmidrule(lr){12-13} \cmidrule(lr){14-15}
 & reward $\uparrow$ & cost $\downarrow$ & reward $\uparrow$ & cost $\downarrow$ & reward $\uparrow$ & cost $\downarrow$ & reward $\uparrow$ & cost $\downarrow$ & reward $\uparrow$ & cost $\downarrow$ & reward $\uparrow$ & cost $\downarrow$ & reward $\uparrow$ & cost $\downarrow$ \\
\midrule
PointButton1 & \textcolor{gray}{0.11} & \textcolor{gray}{1.06} & \textcolor{gray}{0.24} & \textcolor{gray}{1.95} & \textcolor{gray}{0.64} & \textcolor{gray}{3.31} & \textcolor{gray}{0.53} & \textcolor{gray}{2.99} & \textcolor{gray}{0.12} & \textcolor{gray}{1.43} & \textcolor{gray}{0.10} & \textcolor{gray}{1.23} & \textbf{\textcolor{blue}{0.03}} & \textbf{\textcolor{blue}{0.50}} \\
PointButton2 & \textcolor{gray}{0.25} & \textcolor{gray}{2.30} & \textcolor{gray}{0.41} & \textcolor{gray}{2.63} & \textcolor{gray}{0.57} & \textcolor{gray}{4.47} & \textcolor{gray}{0.45} & \textcolor{gray}{2.85} & \textcolor{gray}{0.17} & \textcolor{gray}{1.93} & \textcolor{gray}{0.21} & \textcolor{gray}{1.54} & \textbf{\textcolor{blue}{0.14}} & \textbf{\textcolor{blue}{0.75}} \\
PointCircle1 & \textcolor{gray}{0.81} & \textcolor{gray}{4.97} & \textcolor{gray}{0.69} & \textcolor{gray}{3.04} & \textcolor{gray}{0.45} & \textcolor{gray}{1.14} & \textbf{\textcolor{blue}{0.58}} & \textbf{\textcolor{blue}{0.82}} & \textcolor{gray}{0.85} & \textcolor{gray}{5.39} & \textbf{0.53} & \textbf{0.36} & \textbf{0.50} & \textbf{0.78} \\
PointCircle2 & \textcolor{gray}{0.71} & \textcolor{gray}{5.49} & \textcolor{gray}{0.70} & \textcolor{gray}{4.64} & \textcolor{gray}{0.09} & \textcolor{gray}{4.97} & \textcolor{gray}{0.63} & \textcolor{gray}{1.33} & \textcolor{gray}{0.87} & \textcolor{gray}{8.70} & \textcolor{gray}{0.61} & \textcolor{gray}{1.38} & \textbf{\textcolor{blue}{0.51}} & \textbf{\textcolor{blue}{0.80}} \\
PointGoal1 & \textbf{0.56} & \textbf{0.91} & \textcolor{gray}{0.74} & \textcolor{gray}{1.20} & \textbf{\textcolor{blue}{0.57}} & \textbf{\textcolor{blue}{0.66}} & \textcolor{gray}{0.71} & \textcolor{gray}{1.03} & \textcolor{gray}{0.49} & \textcolor{gray}{1.54} & \textbf{0.17} & \textbf{0.11} & \textbf{0.47} & \textbf{0.53} \\
PointGoal2 & \textcolor{gray}{0.54} & \textcolor{gray}{2.72} & \textcolor{gray}{0.70} & \textcolor{gray}{3.12} & \textcolor{gray}{0.52} & \textcolor{gray}{1.86} & \textcolor{gray}{0.55} & \textcolor{gray}{1.98} & \textcolor{gray}{0.41} & \textcolor{gray}{1.83} & \textbf{\textcolor{blue}{0.20}} & \textbf{\textcolor{blue}{0.55}} & \textbf{\textcolor{blue}{0.33}} & \textbf{\textcolor{blue}{0.80}} \\
PointPush1 & \textbf{0.18} & \textbf{0.77} & \textbf{0.21} & \textbf{0.92} & \textbf{\textcolor{blue}{0.26}} & \textbf{\textcolor{blue}{0.75}} & \textbf{\textcolor{blue}{0.26}} & \textbf{\textcolor{blue}{0.94}} & \textcolor{gray}{0.13} & \textcolor{gray}{1.08} & \textbf{0.19} & \textbf{0.38} & \textbf{0.19} & \textbf{0.29} \\
PointPush2 & \textcolor{gray}{0.15} & \textcolor{gray}{1.39} & \textcolor{gray}{0.17} & \textcolor{gray}{1.16} & \textcolor{gray}{0.09} & \textcolor{gray}{1.57} & \textcolor{gray}{0.21} & \textcolor{gray}{1.13} & \textcolor{gray}{0.04} & \textcolor{gray}{1.17} & \textbf{\textcolor{blue}{0.15}} & \textbf{\textcolor{blue}{0.70}} & \textbf{\textcolor{blue}{0.13}} & \textbf{\textcolor{blue}{0.64}} \\

CarButton1 & \textcolor{gray}{0.00} & \textcolor{gray}{1.54} & \textcolor{gray}{0.22} & \textcolor{gray}{2.76} & \textcolor{gray}{0.42} & \textcolor{gray}{10.20} & \textcolor{gray}{0.21} & \textcolor{gray}{2.71} & \textcolor{gray}{-0.09} & \textcolor{gray}{1.36} & \textcolor{gray}{0.01} & \textcolor{gray}{1.17} & \textbf{\textcolor{blue}{-0.01}} & \textbf{\textcolor{blue}{0.30}} \\
CarButton2 & \textcolor{gray}{-0.03} & \textcolor{gray}{1.38} & \textcolor{gray}{0.00} & \textcolor{gray}{2.00} & \textcolor{gray}{0.36} & \textcolor{gray}{12.65} & \textcolor{gray}{0.19} & \textcolor{gray}{3.53} & \textcolor{gray}{-0.06} & \textcolor{gray}{1.27} & \textcolor{gray}{-0.14} & \textcolor{gray}{1.17} & \textbf{\textcolor{blue}{-0.08}} & \textbf{\textcolor{blue}{0.32}} \\
CarCircle1 & \textcolor{gray}{0.71} & \textcolor{gray}{5.67} & \textcolor{gray}{0.76} & \textcolor{gray}{5.39} & \textcolor{gray}{0.09} & \textcolor{gray}{1.10} & \textcolor{gray}{0.59} & \textcolor{gray}{2.90} & \textcolor{gray}{0.70} & \textcolor{gray}{5.86} & \textcolor{gray}{0.64} & \textcolor{gray}{3.24} & \textcolor{gray}{0.54} & \textcolor{gray}{1.51} \\
CarCircle2 & \textcolor{gray}{0.73} & \textcolor{gray}{7.60} & \textcolor{gray}{0.74} & \textcolor{gray}{6.68} & \textcolor{gray}{0.47} & \textcolor{gray}{2.34} & \textcolor{gray}{0.65} & \textcolor{gray}{4.22} & \textcolor{gray}{0.75} & \textcolor{gray}{8.34} & \textcolor{gray}{0.66} & \textcolor{gray}{4.98} & \textcolor{gray}{0.50} & \textcolor{gray}{1.55} \\
CarGoal1 & \textbf{\textcolor{blue}{0.39}} & \textbf{\textcolor{blue}{0.42}} & \textcolor{gray}{0.60} & \textcolor{gray}{1.10} & \textcolor{gray}{0.80} & \textcolor{gray}{1.51} & \textcolor{gray}{0.67} & \textcolor{gray}{1.18} & \textbf{0.38} & \textbf{0.53} & \textbf{0.28} & \textbf{0.47} & \textbf{0.33} & \textbf{0.38} \\
CarGoal2 & \textbf{\textcolor{blue}{0.23}} & \textbf{\textcolor{blue}{0.78}} & \textcolor{gray}{0.29} & \textcolor{gray}{1.27} & \textcolor{gray}{0.56} & \textcolor{gray}{3.64} & \textcolor{gray}{0.48} & \textcolor{gray}{2.17} & \textbf{0.22} & \textbf{0.82} & \textbf{0.21} & \textbf{0.75} & \textbf{0.16} & \textbf{0.62} \\
CarPush1 & \textbf{0.19} & \textbf{0.39} & \textbf{0.21} & \textbf{0.45} & \textcolor{gray}{0.11} & \textcolor{gray}{1.11} & \textbf{\textcolor{blue}{0.31}} & \textbf{\textcolor{blue}{0.81}} & \textbf{0.16} & \textbf{0.46} & \textbf{0.22} & \textbf{0.34} & \textbf{0.20} & \textbf{0.31} \\
CarPush2 & \textcolor{gray}{0.13} & \textcolor{gray}{1.27} & \textcolor{gray}{0.11} & \textcolor{gray}{1.06} & \textcolor{gray}{0.19} & \textcolor{gray}{4.59} & \textcolor{gray}{0.19} & \textcolor{gray}{1.88} & \textcolor{gray}{0.09} & \textcolor{gray}{1.48} & \textbf{\textcolor{blue}{0.14}} & \textbf{\textcolor{blue}{0.82}} & \textbf{\textcolor{blue}{0.07}} & \textbf{\textcolor{blue}{0.75}} \\

SwimmerVelo & \textcolor{gray}{0.51} & \textcolor{gray}{2.62} & \textcolor{gray}{0.27} & \textcolor{gray}{1.83} & \textcolor{gray}{0.18} & \textcolor{gray}{2.04} & \textbf{\textcolor{blue}{0.63}} & \textbf{\textcolor{blue}{0.88}} & \textcolor{gray}{0.61} & \textcolor{gray}{6.31} & \textbf{0.41} & \textbf{0.51} & \textcolor{gray}{0.43} & \textcolor{gray}{1.58} \\
HopperVelo & \textcolor{gray}{0.54} & \textcolor{gray}{2.80} & \textcolor{gray}{0.36} & \textcolor{gray}{5.47} & \textcolor{gray}{0.12} & \textcolor{gray}{1.73} & \textbf{\textcolor{blue}{0.71}} & \textbf{\textcolor{blue}{0.75}} & \textcolor{gray}{0.10} & \textcolor{gray}{1.45} & \textbf{0.62} & \textbf{0.71} & \textbf{0.41} & \textbf{0.70} \\
HalfCheetahVelo & \textcolor{gray}{0.96} & \textcolor{gray}{5.26} & \textcolor{gray}{0.98} & \textcolor{gray}{6.55} & \textbf{0.34} & \textbf{0.66} & \textbf{\textcolor{blue}{0.98}} & \textbf{\textcolor{blue}{0.16}} & \textbf{0.62} & \textbf{0.00} & \textbf{0.95} & \textbf{0.21} & \textbf{0.94} & \textbf{0.77} \\
Walker2dVelo & \textcolor{gray}{0.74} & \textcolor{gray}{1.85} & \textcolor{gray}{0.89} & \textcolor{gray}{4.16} & \textbf{0.03} & \textbf{0.23} & \textbf{0.79} & \textbf{0.10} & \textbf{0.12} & \textbf{0.86} & \textbf{\textcolor{blue}{0.80}} & \textbf{\textcolor{blue}{0.04}} & \textbf{\textcolor{blue}{0.80}} & \textbf{\textcolor{blue}{0.62}} \\
AntVelo & \textcolor{gray}{0.98} & \textcolor{gray}{2.58} & \textbf{-1.01} & \textbf{0.00} & \textbf{-1.01} & \textbf{0.00} & \textbf{\textcolor{blue}{0.99}} & \textbf{\textcolor{blue}{0.50}} & \textcolor{gray}{1.00} & \textcolor{gray}{2.99} & \textbf{0.89} & \textbf{0.67} & \textbf{0.95} & \textbf{0.64} \\

\midrule
\textbf{SafetyGym Avg} & \textcolor{gray}{0.45} & \textcolor{gray}{2.56} & \textcolor{gray}{0.39} & \textcolor{gray}{2.73} & \textcolor{gray}{0.28} & \textcolor{gray}{2.88} & \textcolor{gray}{0.54} & \textcolor{gray}{1.66} & \textcolor{gray}{0.37} & \textcolor{gray}{2.61} & \textcolor{gray}{0.37} & \textcolor{gray}{1.02} & \textbf{\textcolor{blue}{0.36}} & \textbf{\textcolor{blue}{0.72}} \\

\midrule
BallRun & \textcolor{gray}{0.67} & \textcolor{gray}{3.16} & \textcolor{gray}{0.02} & \textcolor{gray}{3.11} & \textcolor{gray}{0.30} & \textcolor{gray}{2.36} & \textcolor{gray}{0.38} & \textcolor{gray}{1.02} & \textcolor{gray}{0.57} & \textcolor{gray}{3.38} & \textbf{\textcolor{blue}{0.24}} & \textbf{\textcolor{blue}{0.60}} & \textbf{\textcolor{blue}{0.19}} & \textbf{\textcolor{blue}{0.94}} \\
CarRun & \textbf{0.97} & \textbf{0.11} & \textcolor{gray}{0.40} & \textcolor{gray}{7.65} & \textbf{0.93} & \textbf{0.94} & \textbf{\textcolor{blue}{0.98}} & \textbf{\textcolor{blue}{0.63}} & \textbf{0.93} & \textbf{0.00} & \textbf{0.97} & \textbf{0.00} & \textbf{0.97} & \textbf{0.25} \\
DroneRun & \textcolor{gray}{0.32} & \textcolor{gray}{1.44} & \textcolor{gray}{0.42} & \textcolor{gray}{3.85} & \textcolor{gray}{0.36} & \textcolor{gray}{2.62} & \textbf{\textcolor{blue}{0.62}} & \textbf{\textcolor{blue}{0.74}} & \textcolor{gray}{0.67} & \textcolor{gray}{4.27} & \textbf{0.54} & \textbf{0.39} & \textcolor{gray}{0.47} & \textcolor{gray}{2.19} \\
AntRun & \textcolor{gray}{0.72} & \textcolor{gray}{2.76} & \textbf{0.04} & \textbf{0.08} & \textbf{0.02} & \textbf{0.01} & \textbf{\textcolor{blue}{0.72}} & \textbf{\textcolor{blue}{0.82}} & \textbf{0.62} & \textbf{1.00} & \textbf{0.53} & \textbf{0.25} & \textbf{0.61} & \textbf{0.90} \\

BallCircle & \textcolor{gray}{0.76} & \textcolor{gray}{2.70} & \textcolor{gray}{0.80} & \textcolor{gray}{2.22} & \textbf{0.62} & \textbf{0.57} & \textcolor{gray}{0.78} & \textcolor{gray}{1.20} & \textcolor{gray}{0.70} & \textcolor{gray}{2.63} & \textbf{0.59} & \textbf{0.19} & \textbf{\textcolor{blue}{0.69}} & \textbf{\textcolor{blue}{0.59}} \\
CarCircle & \textcolor{gray}{0.62} & \textcolor{gray}{3.15} & \textcolor{gray}{0.76} & \textcolor{gray}{2.78} & \textbf{0.71} & \textbf{0.27} & \textbf{\textcolor{blue}{0.75}} & \textbf{\textcolor{blue}{0.91}} & \textcolor{gray}{0.48} & \textcolor{gray}{2.93} & \textbf{0.61} & \textbf{0.29} & \textbf{0.69} & \textbf{0.65} \\
DroneCircle & \textcolor{gray}{0.71} & \textcolor{gray}{2.70} & \textcolor{gray}{0.80} & \textcolor{gray}{3.93} & \textbf{-0.23} & \textbf{0.36} & \textcolor{gray}{0.62} & \textcolor{gray}{1.02} & \textcolor{gray}{0.25} & \textcolor{gray}{1.02} & \textbf{\textcolor{blue}{0.42}} & \textbf{\textcolor{blue}{0.14}} & \textbf{\textcolor{blue}{0.55}} & \textbf{\textcolor{blue}{0.67}} \\
AntCircle & \textcolor{gray}{0.65} & \textcolor{gray}{4.67} & \textcolor{gray}{0.64} & \textcolor{gray}{5.23} & \textbf{0.00} & \textbf{0.00} & \textcolor{gray}{0.48} & \textcolor{gray}{2.34} & \textcolor{gray}{0.18} & \textcolor{gray}{4.33} & \textbf{\textcolor{blue}{0.47}} & \textbf{\textcolor{blue}{0.42}} & \textbf{0.41} & \textbf{0.15} \\

\midrule
\textbf{BulletGym Avg} & \textcolor{gray}{0.68} & \textcolor{gray}{2.59} & \textcolor{gray}{0.49} & \textcolor{gray}{3.61} & \textbf{0.34} & \textbf{0.89} & \textcolor{gray}{0.67} & \textcolor{gray}{1.09} & \textcolor{gray}{0.55} & \textcolor{gray}{2.45} & \textbf{\textcolor{blue}{0.55}} & \textbf{\textcolor{blue}{0.29}} & \textbf{\textcolor{blue}{0.57}} & \textbf{\textcolor{blue}{0.79}} \\

\midrule
easysparse & \textbf{0.18} & \textbf{0.82} & \textbf{0.03} & \textbf{0.71} & \textbf{-0.06} & \textbf{0.08} & \textbf{\textcolor{blue}{0.25}} & \textbf{\textcolor{blue}{0.26}} & \textcolor{gray}{0.92} & \textcolor{gray}{5.33} & \textcolor{gray}{0.54} & \textcolor{gray}{2.14} & \textbf{0.11} & \textbf{0.34} \\
easymean & \textbf{0.05} & \textbf{0.18} & \textbf{0.02} & \textbf{0.50} & \textbf{-0.06} & \textbf{0.06} & \textcolor{gray}{0.49} & \textcolor{gray}{1.18} & \textcolor{gray}{0.73} & \textcolor{gray}{4.35} & \textbf{\textcolor{blue}{0.16}} & \textbf{\textcolor{blue}{0.41}} & \textbf{0.01} & \textbf{0.20} \\
easydense & \textbf{0.20} & \textbf{0.31} & \textbf{-0.01} & \textbf{0.39} & \textbf{-0.06} & \textbf{0.08} & \textbf{\textcolor{blue}{0.41}} & \textbf{\textcolor{blue}{0.51}} & \textcolor{gray}{0.53} & \textcolor{gray}{3.02} & \textbf{0.11} & \textbf{0.15} & \textbf{0.10} & \textbf{0.19} \\
mediumsparse & \textcolor{gray}{0.65} & \textcolor{gray}{1.56} & \textbf{-0.01} & \textbf{0.31} & \textbf{-0.09} & \textbf{0.06} & \textcolor{gray}{0.55} & \textcolor{gray}{1.03} & \textcolor{gray}{0.79} & \textcolor{gray}{2.65} & \textbf{\textcolor{blue}{0.65}} & \textbf{\textcolor{blue}{0.99}} & \textbf{\textcolor{blue}{0.60}} & \textbf{\textcolor{blue}{0.74}} \\
mediummean & \textcolor{gray}{0.74} & \textcolor{gray}{1.55} & \textbf{-0.04} & \textbf{0.10} & \textbf{-0.08} & \textbf{0.07} & \textcolor{gray}{0.48} & \textcolor{gray}{1.31} & \textcolor{gray}{0.78} & \textcolor{gray}{2.52} & \textbf{\textcolor{blue}{0.13}} & \textbf{\textcolor{blue}{0.23}} & \textbf{\textcolor{blue}{0.66}} & \textbf{\textcolor{blue}{0.94}} \\
mediumdense & \textcolor{gray}{0.63} & \textcolor{gray}{1.08} & \textbf{0.08} & \textbf{0.63} & \textbf{-0.07} & \textbf{0.05} & \textbf{0.21} & \textbf{0.29} & \textcolor{gray}{0.63} & \textcolor{gray}{2.17} & \textbf{\textcolor{blue}{0.69}} & \textbf{\textcolor{blue}{0.80}} & \textbf{\textcolor{blue}{0.69}} & \textbf{\textcolor{blue}{0.56}} \\
hardsparse & \textbf{0.19} & \textbf{0.54} & \textbf{0.02} & \textbf{0.24} & \textbf{-0.04} & \textbf{0.08} & \textbf{0.26} & \textbf{0.63} & \textcolor{gray}{0.38} & \textcolor{gray}{2.31} & \textbf{0.10} & \textbf{0.18} & \textbf{\textcolor{blue}{0.45}} & \textbf{\textcolor{blue}{0.72}} \\
hardmean & \textcolor{gray}{0.47} & \textcolor{gray}{2.32} & \textbf{0.01} & \textbf{0.26} & \textbf{-0.05} & \textbf{0.07} & \textbf{0.19} & \textbf{0.35} & \textcolor{gray}{0.36} & \textcolor{gray}{2.37} & \textbf{0.19} & \textbf{0.30} & \textbf{\textcolor{blue}{0.28}} & \textbf{\textcolor{blue}{0.25}} \\
harddense & \textcolor{gray}{0.36} & \textcolor{gray}{1.62} & \textcolor{gray}{0.07} & \textcolor{gray}{3.86} & \textbf{-0.04} & \textbf{0.05} & \textbf{0.28} & \textbf{0.69} & \textcolor{gray}{0.23} & \textcolor{gray}{1.46} & \textbf{0.22} & \textbf{0.45} & \textbf{\textcolor{blue}{0.37}} & \textbf{\textcolor{blue}{0.67}} \\

\midrule
\textbf{MetaDrive Avg} & \textcolor{gray}{0.39} & \textcolor{gray}{1.11} & \textbf{0.02} & \textbf{0.78} & \textbf{-0.06} & \textbf{0.07} & \textbf{0.35} & \textbf{0.69} & \textcolor{gray}{0.59} & \textcolor{gray}{2.91} & \textbf{\textcolor{blue}{0.31}} & \textbf{\textcolor{blue}{0.63}} & \textbf{\textcolor{blue}{0.36}} & \textbf{\textcolor{blue}{0.51}} \\

\bottomrule
\end{tabular}
\end{adjustbox}
\end{table*}

\vspace{1ex}

\noindent \textbf{Ablation for number of policies.} Tables \ref{tab:iql_heads} and \ref{tab:sac_heads} in the Appendix \ref{ap:heads} show the results of CAPS-$K$ with $K \in \{ 2, 4, 8\}$ heads/policies. We make the following observations. {\bf 1)} CAPS-4 with four policies achieves better results than CAPS-2 with two policies.  CAPS-8 with eight policies despite its potential for increased capacity, does not lead to substantial improvements over CAPS-4. The diminishing returns demonstrate that the incremental benefit of adding more policies is decreasing, with CAPS-4 providing an optimized balance of flexibility and complexity. Eight policies introduce some redundancy and potentially exacerbates stochasticity, rather than providing additional value. {\bf 2)} CAPS-2 with two policies also demonstrates impressive results and is preferred as our main approach due to its simplicity and computational efficiency. Switching between reward-maximizing and cost-minimizing policies is very effective for managing varying test-time cost constraints. In contrast, CAPS variant with more than two heads involve optimizing additional policies for combined reward and cost objectives with extra scalarization hyper-parameters $\lambda_k$. Therefore, we use CAPS with two policies (one for reward maximization and one for cost minimization) for the rest of our experimental analysis.

\vspace{1ex}

\noindent \textbf{Ablation for shared policy representation.} We investigate the advantages of using a shared policy representation architecture compared to training separate agents for reward and cost objectives respectively. A shared representation allows different heads/policies to utilize common state features, to potentially improve the overall performance. The intuition behind this idea is that the shared network enables the cost head to achieve higher rewards and the reward head to incur lower costs compared to independently learned policies.

\begin{figure}[ht]
    \centering
    \includegraphics[width=\columnwidth]{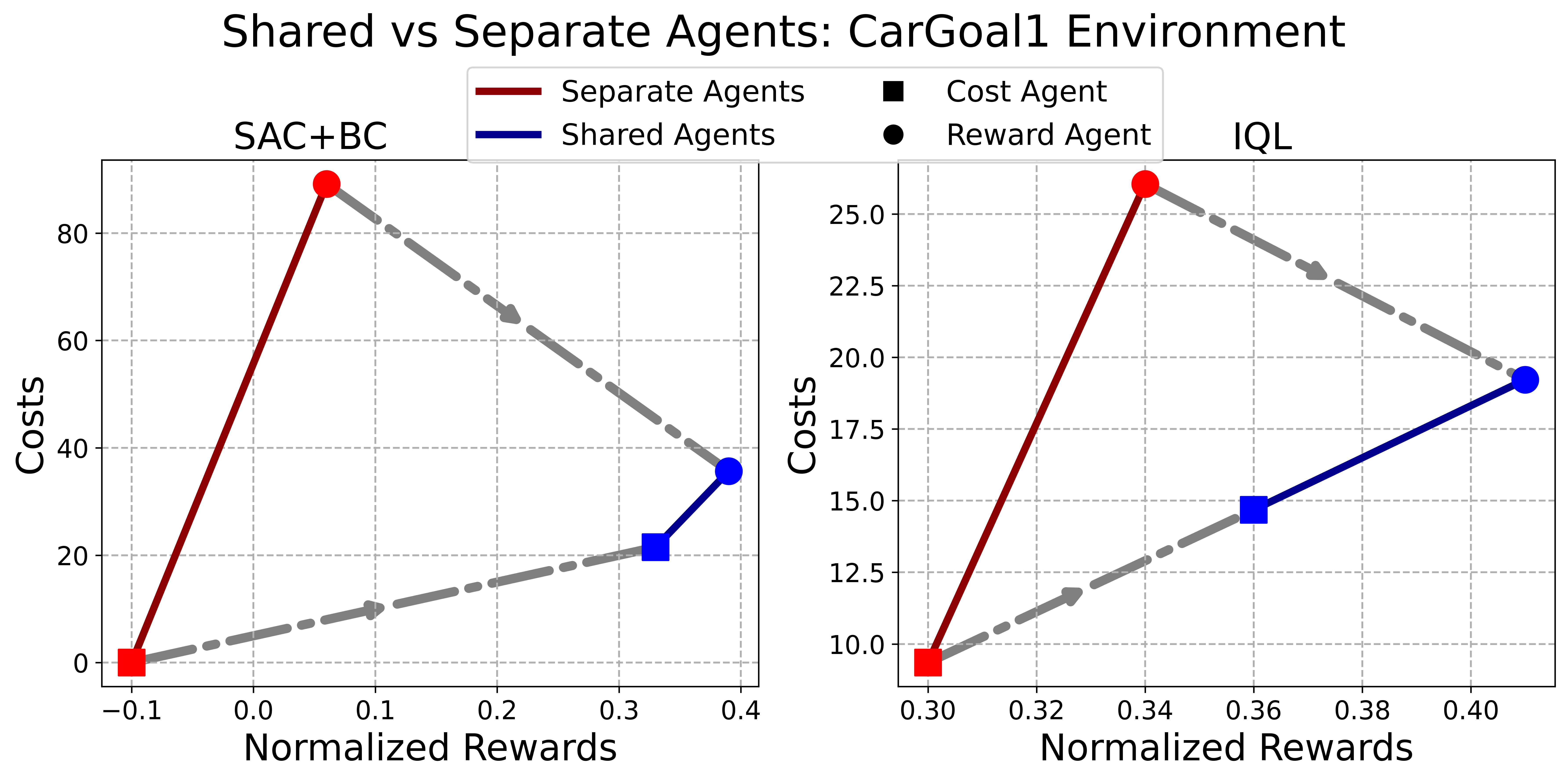}
    \caption{Ablation results for shared architecture vs. independently trained reward and cost optimized policies.}
    \label{fig:sharing}
\end{figure}

For instance, Figure \ref{fig:sharing} compares the performance of separate agents and shared agents in the CarGoal1 environment using the two offline RL instantiations: SAC+BC (left plot) and IQL (right plot). The focus is on understanding how sharing impacts the behavior of cost agents and reward agents when run uniformly.

In the case of separately trained agents, joined by the red line in both plots, the reward agent maximizes rewards without much consideration for the cost objective, resulting in high costs for achieving high rewards. For instance, in the SAC+BC algorithm, the reward agent’s point shows a high cost of around 80 and a moderate reward of about 0.1, indicating a reckless pursuit of rewards. Similarly, in the IQL algorithm, the reward agent incurs a cost of about 25 to achieve a reward of approximately 0.34. Conversely, the cost agent in the separate agents’ scenario aggressively minimizes costs, leading to low rewards as it avoids actions that could increase costs. This conservative behavior is evident the SAC+BC case, where the cost agent achieves low rewards of around -0.1 and costs near 0, and the IQL, where the cost agent’s rewards are around 0.3 with costs around 10.

In contrast, the shared agents, joined by the blue lines, exhibit a more balanced approach. The shared policy representation helps cost agent become less conservative, encouraging them to seek higher rewards even if it means incurring some costs. For example, in the SAC+BC algorithm, the shared cost agent achieves more rewards of about 0.33 with increased costs of around 20, showing a willingness to balance both objectives. Similarly, in the IQL algorithm, the shared cost agent achieves higher rewards of approximately 0.36 with moderate costs of around 15. A similar balanced approach is also evident for the reward agent under sharing. 

Tables \ref{tab:shared_sac_cost}, \ref{tab:shared_sac_reward}, \ref{tab:shared_iql_cost} and \ref{tab:shared_iql_reward} in the Appendix \ref{ap:sharing} shows the results for shared vs. independent policies ablations on all 38 benchmarks for IQL and SAC+BC. Additionally, the performance of CAPS with and without sharing is detailed in Table \ref{tab:shared_switch}. These results demonstrate the benefits of the shared policy representation. Therefore, we use CAPS with shared representation for two policies for baseline comparison.

\noindent \textbf{Comparison with baselines.} 
Table \ref{tab:main_results} compares the two CAPS variants, namely, CAPS(IQL) and CAPS(SAC+BC) against all baseline methods. Each method’s performance is displayed across two objectives: reward maximization and cost minimization. The table includes results for various tasks and aggregated performance for the three environment categories (SafetyGym Avg, BulletGym Avg, MetaDrive Avg). 
We make the following observations. {\bf 1)} Both CAPS instantiations significantly outperform the baselines, meeting cost constraints in 34 out of 38 tasks for CAPS(IQL) and 30 out of 38 tasks for CAPS(SAC+BC). This is remarkable given that safety is the primary metric. In contrast, the second-best method, CDT, only manages to satisfy the constraints in 19 out of 38 tasks. {\bf 2)} CAPS(IQL) achieves the highest reward in 18 out of 38 tasks, demonstrating superior performance in reward maximization while meeting the safety/cost constraints. This result shows that CAPS can leverage better offline RL algorithms to further improve the performance in OSRL. {\bf 3)} Results in Appendix \ref{ap:computation_time} show that the training time of CAPS with increasing number of heads/policies is significantly lower than CDT. Other baseline methods will be worse as they need to be trained for every cost constraint.  {\bf 4)} CAPS with two policies is a minimalist approach and a strong wrapper based baseline for OSRL.

\vspace{1ex}

\noindent \textbf{Ablation on cost limits.} To evaluate the performance of CAPS under various cost limits, we compare both CAPS variants with CDT. 
The results in Table \ref{tab:cost_limits_ablation} demonstrate that CAPS handles different cost constraints better than CDT. 
\begin{table}[ht]
    \centering
    \caption{Performance of CAPS(IQL) and CAPS(SAC+BC) for different cost limit configurations.}
    \resizebox{\columnwidth}{!}{%
    \begin{tabular}{cccc}
        \toprule
        \textbf{Cost Limits} & \textbf{CAPS(IQL)}  & \textbf{CAPS(SAC+BC)}  & \textbf{CDT}\\  
        \midrule
        \{ 5, 10 \}   & \textcolor{blue}{\textbf{18}} / 38 &   \textcolor{blue}{\textbf{15}} / 38&  11 / 38\\ 
        \{10, 20\}  & \textcolor{blue}{\textbf{28}} / 38 &  \textcolor{blue}{\textbf{24}} / 38&  16 / 38\\ 
        \{15, 30\}  & \textcolor{blue}{\textbf{31}} / 38 &  \textcolor{blue}{\textbf{29}} / 38&  19 / 38\\ 
        \{20, 40\}  & \textcolor{blue}{\textbf{33}} / 38 &  \textcolor{blue}{\textbf{32}} / 38&  20 / 38\\ 
        \{30, 60\}  & \textcolor{blue}{\textbf{33}} / 38 &  \textcolor{blue}{\textbf{34}} / 38&  22 / 38\\ 
        \{40, 80\} & \textcolor{blue}{\textbf{33}} / 38 &  \textcolor{blue}{\textbf{35}} / 38&  25 / 38\\ 
        \bottomrule
    \end{tabular}
    }
    \label{tab:cost_limits_ablation}
\end{table}

For a cost limit configuration of \{5, 10\}, where the cost limit is 10 for SafetyGym environments and 5 for the other two categories, CAPS(IQL) maintains safety in 18 out of 38 tasks, while CAPS(SAC+BC) achieves safety in 15 out of 38 tasks (compared to 11 for CDT). Although these numbers might appear modest, they are reasonable given that the benchmark was designed with higher cost thresholds for evaluation.
When the cost limit is increased to \{10, 20\}, both CAPS variants show significant improvement: CAPS(IQL) ensures safety in 28 out of 38 tasks, and CAPS(SAC+BC) in 24 out of 38 tasks. With a configuration of \{15, 30\}, CAPS(IQL) achieves safety in 31 out of 38 tasks, while CAPS(SAC+BC) secures safety in 29 out of 38 tasks.
Performance peaks at the \{20, 40\}, \{30, 60\}, and \{40, 80\} configurations. CAPS(IQL) maintains safety in 33 out of 38 tasks at all three configurations, while CAPS(SAC+BC) shows consistent improvement, achieving safety in 32, 34, and 35 out of 38 tasks, respectively.
The results demonstrates that both CAPS instantiations can effectively adapt to various cost limits without requiring retraining. In contrast to other methods, which often necessitate specific training for each cost limit, both CAPS variants demonstrate flexibility and robustness, consistently outperforming CDT across all cost limit configurations. Full results comparing CDT and CAPS(IQL) can be found in Appendix \ref{ap:cost_lim_ablations}.

\vspace{1ex}

\noindent \textbf{Off-policy evaluation.} We implemented a variant of CAPS with off-policy evaluation for the two policies/heads configuration, comparing this method to our original approach. We trained two Q-functions for each head/policy (i.e., $\pi_r$ and $\pi_c$) using fitted Q-evaluation (FQE) \cite{le2019batch}: $\hat{Q}^r$ for rewards and $\hat{Q}^c$ for costs, applying the same logic for policy switching. In Appendix \ref{ap:fqe}, we present results comparing the original CAPS method to two variants: 1) CAPS with FQE based $\hat{Q}^r$ and $\hat{Q}^c$ for both $\pi_r$ and $\pi_c$ (reward-cost FQE), and 2) CAPS with FQE based $\hat{Q}^r$ for $\pi_r$ and $\pi_c$ (reward FQE). The reward-cost FQE method yields overly conservative results with low rewards. The learned $\hat{Q}^c$ corresponding to respective policies $\pi_r$ and $\pi_c$ frequently estimates that actions exceed the cost limit, defaulting to selecting the action with the lowest cost Q-value. Our results also show that reward FQE method which use $\hat{Q}^r$ functions from FQE and $Q^c$ from offline RL method yields better performance than reward-cost FQE and is comparable to the original CAPS (uses $Q^c$ and $Q^r$ from offline RL). We hypothesize that this behavior is due to estimation errors in off-policy evaluation as noted by prior literature \cite{offline_rl_survey_2}. The FQE based $\hat{Q}^r$ functions effectively differentiate between actions without needing a precise estimation (unlike cost where precision matters to check safety of decisions at a state). These findings support the robust choice of applying $Q^c$ in the filtering stage of CAPS decision-making.

\section{Summary and Future Work}

This paper introduces CAPS, a novel offline safe RL framework that dynamically adapts to varying safety constraints by training multiple diverse policies. CAPS switches between actions from policies during deployment, ensuring safety while maximizing rewards. Empirical results show that CAPS outperforms existing methods. Future work could integrate CAPS into online learning for real-time gains. 

\section*{Acknowledgements} The authors gratefully acknowledge the in part support by USDA-NIFA funded AgAID Institute award 2021-67021-35344. The views expressed are those of the authors and do not reflect the official policy or position of the USDA-NIFA.

\bibliography{aaai25}

\clearpage
\onecolumn

\appendix

\section{Proof of Theorem \ref{thm:bound} for Safety Guarantee of CAPS}

\vspace{2ex}

\setcounter{theorem}{0}
\begin{theorem}
    For any MDP $M$ with optimal-cost variation $\epsilon$ and any $\kappa$-admissible policy $\pi$, we have that for any state $s$, time step $t$, and accumulated cost $c_{<t}$, $$V^{\pi,c}_t(s)\leq \max\{V^c_t(s),\kappa-c_{<t}\} + (T-t)\epsilon$$
\end{theorem}

\noindent 
{\bf Proof:}  
We use proof by induction on the time step $t$ with a base case of $T$. 

\emph{Base Case $(t=T)$:} The base case states that for any $s$ and $c_{<t}$ $V^{\pi,c}_T(s)\leq \max\{V^c_T(s),\kappa-c_{<T}\}$. This follows directly from the facts that for all $s$, $V^{\pi,c}_T(s) = c(s)$ and that $V^c_T(s)=c(s)$.  

\emph{Inductive Step:} Here we assume the theorem holds for time step $t+1$ and show that it follows for time step $t$. Our inductive hypothesis states that for all $s$
\begin{align}
V^{\pi,c}_{t+1}(s)\leq \max\{V^c_{t+1}(s),\kappa-c_{<t+1}\} + (T-t-1)\epsilon    
\label{step:0}
\end{align}
Next expand the value function for $t$,
\begin{small}
\begin{align}
\nonumber & V^{\pi,c}_{t}(s)  = c(s) + \sum_{s'} P(s,\pi_t(s,c_{<t}),s')V^{\pi,c}_{t+1}(s') \\
\nonumber    & \leq c(s) + \sum_{s'} P(s,\pi_t(s,c_{<t}),s')\cdot \max\{V^c_{t+1}(s'),\kappa-c_{<t}-c(s)\} \\
\nonumber    &\hspace{3em}  + (T-t-1)\epsilon \\
\nonumber    & = \sum_{s'} P(s,\pi_t(s,c_{<t}),s')\cdot \max\{V^c_{t+1}(s') + c(s),\kappa-c_{<t}\} \\
    & \hspace{3em}  + (T-t-1)\epsilon    
    \label{step:1}
\end{align}
\end{small}
The first inequality follows from the inductive hypothesis (\ref{step:0}) and the the fact that in the context of $s$, $c_{<t+1}=c_{<t}+c(s)$. The next equality follows by simply pushing $c$ into the expectation noting that $c(s) = \sum_{s'}P(s,a,s')c(s)$ for any $s$ and $a$. 

We now use the assumption that $\pi$ is $\kappa$-admissible, which implies that, 
\begin{align*}
  Q^c_t(s,\pi_t(s,c_{<t}))  & = c(s) + \sum_{s'}  P(s,\pi_t(s,c{<t}),s') V^c_{t+1}(s') \\
    & \leq \max\{V^c_t(s), \kappa-c_{<t}\}
\end{align*}
which we rewrite to
\begin{align*}
 \sum_{s'}  P(s,\pi_t(s,c{<t}),s') & \cdot \left(V^c_{t+1}(s') + c(s)\right) \\
  & \leq \max\{V^c_t(s), \kappa-c_{<t}\}
\end{align*}
We can now combine this with the assumption that our MDP has optimal-cost variation $\epsilon$. Since the sum is an expectation this implies that for any $s'$ 
\begin{align}
    V^c_{t+1}(s') + c(s) \leq \max\{V^c_t(s), \kappa-c_{<t}\} + \epsilon
    \label{step:2}
\end{align} 
By combining (\ref{step:1}) and (\ref{step:2}) we conclude 
\begin{small}
\begin{align*}
V^{\pi,c}_{t}(s)  & \leq \sum_{s'} P(s,\pi_t(s,c_{<t}),s')\cdot \max\{V^c_{t}(s),\kappa-c_{<t}\} + \epsilon    \\ 
    & \hspace{3em}  + (T-t-1)\epsilon  \\
    & = \max\{V^c_{t}(s),\kappa-c_{<t}\} + (T-t)\epsilon 
\end{align*}
\end{small}
which completes the proof. \hfill$\square$

\subsection{Discussion on Assumptions for Safety Guarantees of CAPS}

\vspace{1ex}

{Assumptions 1 and 2 are essential for achieving safety guarantees under CAPS. Intuitively, Assumption 1 ensures that there exists at least one safe action at any intermediate step such that future costs can satisfy the cost constraint based on already observed costs. Having more heads theoretically ensures better coverage of the action space for each agent. Without such a safe action, no algorithm can guarantee safety. We include this assumption to generalize the theorem, making it applicable to other policies or strategies that meet the same condition, not just CAPS. This broadens the applicability of our theoretical analysis.}

\vspace{1ex}

{Assumption 2 helps define the types of Markov Decision Processes (MDPs) for which we can ensure safety guarantees. Without this assumption, we can easily create pathological counterexamples where guarantees would fail. In practice, most of our experimental benchmark environments comply with Assumption 2 with only a small epsilon value. This reflects realistic conditions where good policies demonstrate some degree of recoverability and avoid extreme scenarios such as falling into irrecoverable states or encountering unexpected high rewards.}

\vspace{1ex}

{These assumptions help frame the theoretical results, ensuring that CAPS is validated under reasonable conditions applicable to our benchmark environments.}

\section{Ablation Results for CAPS}
\vspace{2ex}
\subsection{Results of CAPS with Varying Number of Heads} \label{ap:heads}
Table \ref{tab:iql_heads} compares the performance of the CAPS(IQL) instantiation with varying numbers of heads: 2, 4, and 8.  As the number of heads increases, there is a slight improvement in rewards, but this comes with increased complexity and some potential risk. The 2-head configuration is the simplest and safest, consistently maintaining safety across most tasks while delivering reasonable rewards. The 4-head configuration strikes a good balance, offering slightly better rewards than the 2-head setup while still maintaining safety. In contrast, the 8-head configuration, although it achieves the highest rewards in some tasks, introduces greater complexity and sometimes exceeds the safety threshold, marking certain agents as unsafe. For example, in the “PointButton2” task, while the reward remains constant at 0.14 across all configurations, the cost exceeds the safety threshold (1.01) only in the 8-head configuration, highlighting the increased risk associated with more complex setups.
\begin{table*}[ht!]
\centering
\caption{Complete evaluation results of normalized reward and cost across different head configurations (2 Heads, 4 Heads, 8 Heads) for the CAPS (IQL) instantiation. The cost threshold is 1. The $\uparrow$ symbol denotes that higher reward is better. The $\downarrow$ symbol denotes that lower cost (up to threshold 1) is better. Each value is averaged over 3 distinct cost thresholds, 20 evaluation episodes, and 3 random seeds. {\bf Bold:} Safe agents whose normalized cost is smaller than 1. \textcolor{gray}{Gray}: Unsafe agents.}
\label{tab:iql_heads}
\begin{adjustbox}{max width=\textwidth}
\begin{tabular}{lcccccccccccccc}
\toprule
CAPS(IQL) & \multicolumn{2}{c}{2 Heads} & \multicolumn{2}{c}{4 Heads} & \multicolumn{2}{c}{8 Heads} \\
\cmidrule(lr){2-3} \cmidrule(lr){4-5} \cmidrule(lr){6-7} 
 Task & reward $\uparrow$ & cost $\downarrow$ & reward $\uparrow$ & cost $\downarrow$ & reward $\uparrow$ & cost $\downarrow$ \\
\midrule
PointButton1 & \textbf{0.03} & \textbf{0.5} & \textbf{0.06} & \textbf{0.5} & \textbf{0.08} & \textbf{0.57} \\
PointButton2 & \textbf{0.14} & \textbf{0.75} & \textbf{0.14} & \textbf{0.82} & \textbf{0.14} & \textcolor{gray}{1.01} \\
PointCircle1 & \textbf{0.5} & \textbf{0.78} & \textbf{0.52} & \textbf{0.7} & \textbf{0.5} & \textbf{0.76} \\
PointCircle2 & \textbf{0.51} & \textbf{0.8} & \textbf{0.51} & \textbf{0.7} & \textbf{0.5} & \textbf{0.63} \\
PointGoal1 & \textbf{0.47} & \textbf{0.53} & \textbf{0.5} & \textbf{0.6} & \textbf{0.45} & \textbf{0.48} \\
PointGoal2 & \textbf{0.33} & \textbf{0.8} & \textbf{0.35} & \textbf{0.83} & \textbf{0.36} & \textbf{0.85} \\
PointPush1 & \textbf{0.19} & \textbf{0.29} & \textbf{0.17} & \textbf{0.29} & \textbf{0.15} & \textbf{0.46} \\
PointPush2 & \textbf{0.13} & \textbf{0.64} & \textbf{0.13} & \textbf{0.57} & \textbf{0.12} & \textbf{0.62} \\

CarButton1 & \textbf{-0.01} & \textbf{0.3} & \textbf{0} & \textbf{0.47} & \textbf{-0.01} & \textbf{0.25} \\
CarButton2 & \textbf{-0.08} & \textbf{0.32} & \textbf{-0.06} & \textbf{0.33} & \textbf{-0.17} & \textbf{0.35} \\
CarCircle1 & \textcolor{gray}{0.54} & \textcolor{gray}{1.51} & \textcolor{gray}{0.56} & \textcolor{gray}{1.64} & \textcolor{gray}{0.52} & \textcolor{gray}{1.41} \\
CarCircle2 & \textcolor{gray}{0.5} & \textcolor{gray}{1.55} & \textbf{0.43} & \textbf{0.96} & \textcolor{gray}{0.52} & \textcolor{gray}{2} \\
CarGoal1 & \textbf{0.33} & \textbf{0.38} & \textbf{0.35} & \textbf{0.42} & \textbf{0.35} & \textbf{0.45} \\
CarGoal2 & \textbf{0.16} & \textbf{0.62} & \textbf{0.19} & \textbf{0.73} & \textbf{0.16} & \textbf{0.6} \\
CarPush1 & \textbf{0.2} & \textbf{0.31} & \textbf{0.17} & \textbf{0.35} & \textbf{0.18} & \textbf{0.27} \\
CarPush2 & \textbf{0.07} & \textbf{0.75} & \textbf{0.07} & \textbf{0.65} & \textbf{0.06} & \textbf{0.84} \\

SwimmerVelocity & \textcolor{gray}{0.43} & \textcolor{gray}{1.58} & \textcolor{gray}{0.38} & \textcolor{gray}{2.62} & \textbf{0.42} & \textbf{0.97} \\
HopperVelocity & \textbf{0.41} & \textbf{0.7} & \textbf{0.26} & \textbf{0.46} & \textbf{0.35} & \textbf{0.5} \\
HalfCheetahVelocity & \textbf{0.94} & \textbf{0.77} & \textbf{0.95} & \textbf{0.79} & \textbf{0.95} & \textbf{0.73} \\
Walker2dVelocity & \textbf{0.8} & \textbf{0.62} & \textbf{0.79} & \textbf{0.61} & \textbf{0.8} & \textbf{0.74} \\
AntVelocity & \textbf{0.95} & \textbf{0.64} & \textbf{0.96} & \textbf{0.5} & \textbf{0.97} & \textbf{0.54} \\

\midrule
\textbf{SafetyGym Avg} & \textbf{0.36} & \textbf{0.72} & \textbf{0.35} & \textbf{0.74} & \textbf{0.35} & \textbf{0.72} \\
\midrule

BallRun & \textbf{0.19} & \textbf{0.94} & \textbf{0.26} & \textbf{0.77} & \textcolor{gray}{0.23} & \textcolor{gray}{1.13} \\
CarRun & \textbf{0.97} & \textbf{0.25} & \textbf{0.98} & \textbf{0.83} & \textbf{0.97} & \textbf{0.26} \\
DroneRun & \textcolor{gray}{0.47} & \textcolor{gray}{2.19} & \textcolor{gray}{0.5} & \textcolor{gray}{1.55} & \textcolor{gray}{0.47} & \textcolor{gray}{1.56} \\
AntRun & \textbf{0.61} & \textbf{0.9} & \textbf{0.61} & \textbf{0.82} & \textbf{0.62} & \textbf{0.79} \\
BallCircle & \textbf{0.69} & \textbf{0.59} & \textbf{0.7} & \textbf{0.63} & \textbf{0.7} & \textbf{0.64} \\
CarCircle & \textbf{0.69} & \textbf{0.65} & \textbf{0.67} & \textbf{0.68} & \textbf{0.68} & \textbf{0.71} \\
DroneCircle & \textbf{0.55} & \textbf{0.67} & \textbf{0.57} & \textbf{0.7} & \textbf{0.57} & \textbf{0.7} \\
AntCircle & \textbf{0.41} & \textbf{0.15} & \textbf{0.42} & \textbf{0.15} & \textbf{0.42} & \textbf{0.14} \\

\midrule
\textbf{BulletGym Avg} & \textbf{0.57} & \textbf{0.79} & \textbf{0.59} & \textbf{0.77} & \textbf{0.58} & \textbf{0.74} \\
\midrule

easysparse & \textbf{0.11} & \textbf{0.34} & \textbf{0.16} & \textbf{0.21} & \textcolor{gray}{0.32} & \textcolor{gray}{1.58} \\
easymean & \textbf{0.01} & \textbf{0.2} & \textbf{0.23} & \textbf{0.5} & \textbf{0.03} & \textbf{0.09} \\
easydense & \textbf{0.1} & \textbf{0.19} & \textbf{0.22} & \textbf{0.51} & \textbf{0.18} & \textbf{0.53} \\
mediumsparse & \textbf{0.6} & \textbf{0.74} & \textbf{0.56} & \textbf{1} & \textbf{0.61} & \textbf{0.28} \\
mediummean & \textbf{0.66} & \textbf{0.94} & \textbf{0.77} & \textbf{0.47} & \textbf{0.44} & \textbf{0.47} \\
mediumdense & \textbf{0.69} & \textbf{0.56} & \textbf{0.27} & \textbf{0.29} & \textbf{0.29} & \textbf{0.3} \\
hardsparse & \textbf{0.45} & \textbf{0.72} & \textbf{0.34} & \textbf{0.74} & \textbf{0.24} & \textbf{0.85} \\
hardmean & \textbf{0.28} & \textbf{0.25} & \textbf{0.34} & \textbf{0.88} & \textbf{0.3} & \textbf{0.46} \\
harddense & \textbf{0.37} & \textbf{0.67} & \textbf{0.33} & \textbf{0.55} & \textbf{0.25} & \textbf{0.42} \\

\midrule
\textbf{MetaDrive Avg} & \textbf{0.40} & \textbf{0.54} & \textbf{0.36} & \textbf{0.57} & \textbf{0.29} & \textbf{0.43} \\
\bottomrule
\end{tabular}
\end{adjustbox}
\end{table*}

\vspace{1ex}

Table \ref{tab:sac_heads} compares the performance of the CAPS(SAC+BC) method across 2, 4, and 8-head configurations. The results indicate that while increasing the number of heads can lead to slight improvements in rewards, it often results in variations in cost, with some configurations occasionally exceeding the safety threshold.

\vspace{1ex}

The 2-head variant consistently achieves a strong balance between reward and safety, keeping costs well within the acceptable limits. Furthermore, the 2-head setup is less computationally demanding, making it more efficient and faster to train and deploy, solidifying it as a strong choice free from hyper-parameters.

\begin{table*}[ht!]
\centering
\caption{Complete evaluation results of normalized reward and cost across different head configurations (2 Heads, 4 Heads, 8 Heads) for the CAPS (SAC+BC) instantiation. The cost threshold is 1. The $\uparrow$ symbol denotes that higher reward is better. The $\downarrow$ symbol denotes that lower cost (up to threshold 1) is better. Each value is averaged over 3 distinct cost thresholds, 20 evaluation episodes, and 3 random seeds. \textbf{Bold}: Safe agents whose normalized cost $\leq$ 1. \textcolor{gray}{Gray}: Unsafe agents.}
\label{tab:sac_heads}
\begin{adjustbox}{max width=\textwidth}
\begin{tabular}{lcccccccccccccc}
\toprule
CAPS(SAC+BC) & \multicolumn{2}{c}{2 Heads} & \multicolumn{2}{c}{4 Heads} & \multicolumn{2}{c}{8 Heads} \\
\cmidrule(lr){2-3} \cmidrule(lr){4-5} \cmidrule(lr){6-7} 
 Task & reward $\uparrow$ & cost $\downarrow$ & reward $\uparrow$ & cost $\downarrow$ & reward $\uparrow$ & cost $\downarrow$ \\
\midrule
PointButton1 & \textcolor{gray}{0.1} & \textcolor{gray}{1.23} & \textbf{0.06} & \textbf{0.81} & \textbf{0.05} & \textbf{0.75} \\
PointButton2 & \textcolor{gray}{0.21} & \textcolor{gray}{1.54} & \textbf{0.15} & \textbf{0.90} & \textbf{0.15} & \textbf{0.87} \\
PointCircle1 & \textbf{0.53} & \textbf{0.36} & \textbf{0.56} & \textbf{0.46} & \textbf{0.56} & \textbf{0.58} \\
PointCircle2 & \textcolor{gray}{0.61} & \textcolor{gray}{1.38} & \textcolor{gray}{0.61} & \textcolor{gray}{1.35} & \textbf{0.55} & \textbf{0.51} \\
PointGoal1 & \textbf{0.17} & \textbf{0.11} & \textbf{0.11} & \textbf{0.06} & \textbf{0.07} & \textbf{0.06} \\
PointGoal2 & \textbf{0.20} & \textbf{0.55} & \textbf{0.18} & \textbf{0.47} & \textbf{0.15} & \textbf{0.46} \\
PointPush1 & \textbf{0.19} & \textbf{0.38} & \textbf{0.15} & \textbf{0.47} & \textbf{0.17} & \textbf{0.37} \\
PointPush2 & \textbf{0.15} & \textbf{0.70} & \textbf{0.14} & \textbf{1.00} & \textbf{0.16} & \textbf{0.76} \\

CarButton1 & \textcolor{gray}{0.01} & \textcolor{gray}{1.17} & \textbf{0.04} & \textbf{0.97} & \textbf{0.03} & \textbf{0.62} \\
CarButton2 & \textcolor{gray}{-0.14} & \textcolor{gray}{1.17} & \textbf{-0.14} & \textbf{0.92} & \textbf{-0.15} & \textbf{0.97} \\
CarCircle1 & \textcolor{gray}{0.64} & \textcolor{gray}{3.24} & \textcolor{gray}{0.66} & \textcolor{gray}{3.64} & \textcolor{gray}{0.66} & \textcolor{gray}{3.81} \\
CarCircle2 & \textcolor{gray}{0.66} & \textcolor{gray}{4.98} & \textcolor{gray}{0.62} & \textcolor{gray}{4.32} & \textcolor{gray}{0.65} & \textcolor{gray}{4.68} \\
CarGoal1 & \textbf{0.28} & \textbf{0.47} & \textbf{0.35} & \textbf{0.50} & \textbf{0.23} & \textbf{0.37} \\
CarGoal2 & \textbf{0.21} & \textbf{0.75} & \textbf{0.18} & \textbf{0.67} & \textbf{0.15} & \textbf{0.52} \\
CarPush1 & \textbf{0.22} & \textbf{0.34} & \textbf{0.21} & \textbf{0.61} & \textbf{0.22} & \textbf{0.60} \\
CarPush2 & \textbf{0.14} & \textbf{0.82} & \textcolor{gray}{0.09} & \textcolor{gray}{1.09} & \textbf{0.06} & \textbf{0.97} \\

SwimmerVelocity & \textbf{0.41} & \textbf{0.51} & \textbf{0.40} & \textbf{0.13} & \textbf{0.39} & \textbf{0.22} \\
HopperVelocity & \textbf{0.62} & \textbf{0.71} & \textbf{0.80} & \textbf{0.32} & \textbf{0.40} & \textbf{0.45} \\
HalfCheetahVelocity & \textbf{0.95} & \textbf{0.21} & \textbf{0.95} & \textbf{0.35} & \textbf{0.95} & \textbf{0.20} \\
Walker2dVelocity & \textbf{0.80} & \textbf{0.04} & \textbf{0.73} & \textbf{0.00} & \textbf{0.80} & \textbf{0.26}\\
AntVelocity & \textbf{0.89} & \textbf{0.67} & \textbf{0.95} & \textbf{0.77} & \textbf{0.96} & \textbf{0.73} \\

\midrule
\textbf{SafetyGym Avg} & \textcolor{gray}{0.37} & \textcolor{gray}{1.02} & \textbf{0.37} & \textbf{0.94} & \textbf{0.34}& \textbf{0.89}\\
\midrule

BallRun & \textbf{0.24} & \textbf{0.60} & \textbf{0.25} & \textbf{0.41} & \textbf{0.16} & \textbf{0.38} \\
CarRun & \textbf{0.97} & \textbf{0.00} & \textbf{0.97} & \textbf{0.00} & \textbf{0.96} & \textbf{0.00} \\
DroneRun & \textbf{0.54} & \textbf{0.39} & \textbf{0.57} & \textbf{0.30} & \textbf{0.58} & \textbf{0.57} \\
AntRun & \textbf{0.53} & \textbf{0.25} & \textbf{0.38} & \textbf{0.03} & \textbf{0.26} & \textbf{0.12} \\
BallCircle & \textbf{0.59} & \textbf{0.19} & \textbf{0.68} & \textbf{0.31} & \textbf{0.69} & \textbf{0.35} \\
CarCircle & \textbf{0.61} & \textbf{0.29} & \textbf{0.67} & \textbf{0.42} & \textbf{0.63} & \textbf{0.80} \\
DroneCircle & \textbf{0.42} & \textbf{0.14} & \textbf{0.50} & \textbf{0.30} & \textbf{0.51} & \textbf{0.33} \\
AntCircle & \textbf{0.47} & \textbf{0.42} & \textbf{0.42} & \textbf{0.02} & \textbf{0.47} & \textbf{0.06} \\

\midrule
\textbf{BulletGym Avg} & \textbf{0.55} & \textbf{0.29} & \textbf{0.56} & \textbf{0.22} & \textbf{0.53} & \textbf{0.32} \\
\midrule

Easysparse & \textcolor{gray}{0.54} & \textcolor{gray}{2.14} & \textbf{0.09} & \textbf{0.58} & \textbf{0.09} & \textbf{0.49} \\
Easymean & \textbf{0.16} & \textbf{0.41} & \textbf{0.16} & \textbf{0.44} & \textcolor{gray}{0.47} & \textcolor{gray}{1.94} \\
Easydense & \textbf{0.11} & \textbf{0.15} & \textbf{0.20} & \textbf{0.71} & \textcolor{gray}{0.26} & \textcolor{gray}{1.24} 	\\
mediumsparse & \textbf{0.65} & \textbf{0.99} & \textbf{0.50} & \textbf{0.87} & \textbf{0.37} & \textbf{0.65} \\
mediummean & \textbf{0.13} & \textbf{0.23} & \textcolor{gray}{0.59} & \textcolor{gray}{1.47} & \textcolor{gray}{0.80} & \textcolor{gray}{1.66} \\ 
mediumdense & \textbf{0.69} & \textbf{0.80} & \textcolor{gray}{0.59} & \textcolor{gray}{1.23} & \textbf{0.41} & \textbf{0.62} \\
hardsparse & \textbf{0.10} & \textbf{0.18} & \textbf{0.13} & \textbf{0.25} & \textbf{0.30} & \textbf{0.48} \\
hardmean & \textbf{0.19} & \textbf{0.30} & \textbf{0.11} & \textbf{0.34} & \textbf{0.25} & \textbf{0.56} \\
harddense & \textbf{0.22} & \textbf{0.45} & \textbf{0.16} & \textbf{0.27} & \textbf{0.15} & \textbf{0.24} \\

\midrule
\textbf{MetaDrive Avg} & \textbf{0.31} & \textbf{0.63} & \textbf{0.28} & \textbf{0.68} & \textbf{0.34} & \textbf{0.88}  \\
\bottomrule
\end{tabular}
\end{adjustbox}
\end{table*}

\subsection{Results for Shared Architecture Ablation}\label{ap:sharing}

\vspace{1ex}

\subsubsection{CAPS with shared backbone agents vs. separate agents.}
Table \ref{tab:shared_switch} compares the performance of SAC+BC and IQL instantiations of CAPS using separate versus shared backbones across various tasks. The results highlight the significant benefits of using a shared backbone, particularly for SAC+BC variant.

\vspace{1ex}

\noindent\texttt{CAPS(SAC+BC).} SAC+BC agents with separate backbones often struggle, as seen in tasks such as CarGoal1 and AntVelocity. In CarGoal1, the reward jumps from 0.00 with separate agents to 0.28 with a shared backbone, while costs are well managed in both cases. Similarly, in AntVelocity, the reward increases dramatically from -0.78 with separate agents to 0.89 with a shared backbone, highlighting how separate backbones lead to poor performance. This is due to SAC’s training process, where separate agents result in different Q-function updates, degrading the performance. However, sharing the backbone stabilizes these updates and significantly improves overall decisions.

\vspace{1ex}

\noindent\texttt{CAPS(IQL).} For IQL, the impact of backbone sharing is less pronounced but still positive. Since IQL doesn’t query actions outside the dataset during training, the Q-functions are the same whether agents are shared or not. For example, in CarGoal2, the reward slightly improves from 0.15 to 0.16 when using a shared backbone. Although the gains are modest, they suggest that even IQL agents benefit from the consistency of a shared backbone.

\vspace{1.0ex}

Sharing the backbone is crucial for SAC+BC agents, as it prevents performance degradation seen with separate agents. For IQL agents, while the improvements are smaller, shared backbones still offer a slight edge in performance. These findings emphasize the importance of considering backbone sharing to enhance agent performance, particularly in algorithms that rely on querying out-of-distribution actions.

\begin{table*}[ht!]
\centering
\caption{Comparison of evaluation results for normalized reward and cost across different environments of SAC+BC and IQL agents, examining the impact of using separate versus shared backbones. The cost threshold is set to 1. The $\uparrow$ symbol indicates that a higher reward is better, while the $\downarrow$ symbol indicates that a lower cost (up to the threshold of 1) is better. Each value is averaged over 3 distinct cost thresholds, 20 evaluation episodes, and 3 random seeds.}
\label{tab:shared_switch}
\begin{adjustbox}{max width=\textwidth}
\begin{tabular}{lcccccccccccccccc}
\toprule
CAPS & \multicolumn{2}{c}{SAC+BC Seperate Agents} & \multicolumn{2}{c}{SAC+BC Shared Backbone} & \multicolumn{2}{c}{IQL Seperate Agents} & \multicolumn{2}{c}{IQL Shared Backbone} \\
\cmidrule(lr){2-3} \cmidrule(lr){4-5} \cmidrule(lr){6-7} \cmidrule(lr){7-8} 
 Task & reward $\uparrow$ & cost $\downarrow$ & reward $\uparrow$ & cost $\downarrow$ & reward $\uparrow$ & cost $\downarrow$ & reward $\uparrow$ & cost $\downarrow$  \\
\midrule
PointButton1 & \textbf{-0.10} & \textbf{0.15} & \textcolor{gray}{0.10} & \textcolor{gray}{1.23} & \textbf{0.02} & \textbf{0.49} & \textbf{0.03} & \textbf{0.50} \\
PointButton2 & \textbf{-0.03} & \textbf{0.46} & \textcolor{gray}{0.21} & \textcolor{gray}{1.54} & \textbf{0.14} & \textbf{0.97} & \textbf{0.14} & \textbf{0.75} \\
PointCircle1 & \textbf{0.49} & \textbf{0.97} & \textbf{0.53} & \textbf{0.36} & \textcolor{gray}{0.43} & \textcolor{gray}{1.32} & \textbf{0.50} & \textbf{0.78} \\
PointCircle2 & \textbf{0.14} & \textbf{0.34} & \textcolor{gray}{0.61} & \textcolor{gray}{1.38} & \textbf{0.51} & \textbf{0.80} & \textbf{0.51} & \textbf{0.80} \\
PointGoal1 & \textbf{0.02} & \textbf{0.31} & \textbf{0.17} & \textbf{0.11} & \textbf{0.37} & \textbf{0.45} & \textbf{0.47} & \textbf{0.53} \\
PointGoal2 & \textbf{-0.12} & \textbf{0.52} & \textbf{0.20} & \textbf{0.55} & \textbf{0.25} & \textbf{0.50} & \textbf{0.33} & \textbf{0.80} \\
PointPush1 & \textbf{-0.07} & \textbf{0.07} & \textbf{0.19} & \textbf{0.38} & \textbf{0.15} & \textbf{0.38} & \textbf{0.19} & \textbf{0.29} \\
PointPush2 & \textbf{-0.13} & \textbf{0.05} & \textbf{0.15} & \textbf{0.70} & \textbf{0.10} & \textbf{0.58} & \textbf{0.13} & \textbf{0.64} \\

CarButton1 & \textbf{-0.09} & \textbf{0.36} & \textcolor{gray}{0.01} & \textcolor{gray}{1.17} & \textbf{-0.03} & \textbf{0.19} & \textbf{-0.01} & \textbf{0.30} \\
CarButton2 & \textbf{-0.05} & \textbf{0.19} & \textcolor{gray}{-0.14} & \textcolor{gray}{1.17} & \textbf{-0.08} & \textbf{0.35} & \textbf{-0.08} & \textbf{0.32} \\
CarCircle1 & \textcolor{gray}{-0.08} & \textcolor{gray}{1.16} & \textcolor{gray}{0.64} & \textcolor{gray}{3.24} & \textcolor{gray}{0.55} & \textcolor{gray}{1.62} & \textcolor{gray}{0.54} & \textcolor{gray}{1.51} \\
CarCircle2 & \textbf{0.24} & \textbf{0.20} & \textcolor{gray}{0.66} & \textcolor{gray}{4.98} & \textcolor{gray}{0.49} & \textcolor{gray}{1.62} & \textcolor{gray}{0.50} & \textcolor{gray}{1.55} \\
CarGoal1 & \textbf{0.00} & \textbf{0.20} & \textbf{0.28} & \textbf{0.47} & \textbf{0.31} & \textbf{0.40} & \textbf{0.33} & \textbf{0.38} \\
CarGoal2 & \textbf{0.00} & \textbf{0.48} & \textbf{0.21} & \textbf{0.75} & \textbf{0.15} & \textbf{0.60} & \textbf{0.16} & \textbf{0.62} \\
CarPush1 & \textbf{-0.20} & \textbf{0.02} & \textbf{0.22} & \textbf{0.34} & \textbf{0.19} & \textbf{0.36} & \textbf{0.20} & \textbf{0.31} \\
CarPush2 & \textbf{-0.13} & \textbf{0.04} & \textbf{0.14} & \textbf{0.82} & \textbf{0.09} & \textbf{0.65} & \textbf{0.07} & \textbf{0.75} \\

SwimmerVelocity & \textbf{0.13} & \textbf{0.81} & \textbf{0.41} & \textbf{0.51} & \textcolor{gray}{0.43} & \textcolor{gray}{2.08} & \textcolor{gray}{0.46} & \textcolor{gray}{1.95} \\
HopperVelocity & \textbf{0.03} & \textbf{0.44} & \textbf{0.62} & \textbf{0.71} & \textbf{0.37} & \textbf{0.83} & \textbf{0.46} & \textbf{0.80} \\
HalfCheetahVelocity & \textbf{0.73} & \textbf{0.99} & \textbf{0.95} & \textbf{0.21} & \textbf{0.93} & \textbf{0.63} & \textbf{0.94} & \textbf{0.78} \\
Walker2dVelocity & \textbf{0.05} & \textbf{0.30} & \textbf{0.80} & \textbf{0.04} & \textbf{0.75} & \textbf{0.70} & \textbf{0.79} & \textbf{0.68} \\
AntVelocity & \textbf{-0.78} & \textbf{0.00} & \textbf{0.89} & \textbf{0.67} & \textbf{0.94} & \textbf{0.71} & \textbf{0.95} & \textbf{0.62} \\

\midrule
\textbf{SafetyGym Avg} & \textbf{0.00} & \textbf{0.38} & \textcolor{gray}{0.37} & \textcolor{gray}{1.02} & \textbf{0.34} & \textbf{0.77} & \textbf{0.36} & \textbf{0.75} \\

\midrule
BallRun & \textcolor{gray}{0.23} & \textcolor{gray}{1.77} & \textbf{0.24} & \textbf{0.60} & \textbf{0.16} & \textbf{0.90} & \textbf{0.19} & \textbf{0.94} \\
CarRun & \textbf{0.60} & \textbf{0.99} & \textbf{0.97} & \textbf{0.00} & \textbf{0.95} & \textbf{0.12} & \textbf{0.97} & \textbf{0.25} \\
DroneRun & \textcolor{gray}{0.46} & \textcolor{gray}{3.49} & \textbf{0.54} & \textbf{0.39} & \textcolor{gray}{0.47} & \textcolor{gray}{2.47} & \textcolor{gray}{0.47} & \textcolor{gray}{2.19} \\
AntRun & \textbf{0.07} & \textbf{0.02} & \textbf{0.53} & \textbf{0.25} & \textbf{0.61} & \textbf{0.85} & \textbf{0.61} & \textbf{0.90} \\
BallCircle & \textcolor{gray}{0.53} & \textcolor{gray}{1.11} & \textbf{0.59} & \textbf{0.19} & \textbf{0.67} & \textbf{0.72} & \textbf{0.69} & \textbf{0.59} \\
CarCircle & \textcolor{gray}{0.28} & \textcolor{gray}{1.83} & \textbf{0.61} & \textbf{0.29} & \textbf{0.70} & \textbf{0.70} & \textbf{0.69} & \textbf{0.65} \\
DroneCircle & \textbf{-0.21} & \textbf{0.67} & \textbf{0.42} & \textbf{0.14} & \textbf{0.55} & \textbf{0.76} & \textbf{0.55} & \textbf{0.67} \\
AntCircle & \textbf{0.01} & \textbf{0.04} & \textbf{0.47} & \textbf{0.42} & \textbf{0.39} & \textbf{0.19} & \textbf{0.41} & \textbf{0.15} \\

\midrule
\textbf{BulletGym Avg} & \textcolor{gray}{0.25} & \textcolor{gray}{1.24} & \textbf{0.55} & \textbf{0.29} & \textbf{0.56} & \textbf{0.84} & \textbf{0.57} & \textbf{0.79} \\

\midrule
easysparse & \textbf{-0.04} & \textbf{0.17} & \textcolor{gray}{0.54} & \textcolor{gray}{2.14} & \textbf{0.11} & \textbf{0.30} & \textbf{0.11} & \textbf{0.34} \\
easymean & \textbf{-0.06} & \textbf{0.08} & \textbf{0.16} & \textbf{0.41} & \textbf{0.01} & \textbf{0.20} & \textbf{0.01} & \textbf{0.20} \\
easydense & \textbf{-0.05} & \textbf{0.10} & \textbf{0.11} & \textbf{0.15} & \textbf{0.11} & \textbf{0.19} & \textbf{0.10} & \textbf{0.19} \\
mediumsparse & \textbf{-0.08} & \textbf{0.07} & \textbf{0.65} & \textbf{0.99} & \textbf{0.59} & \textbf{0.85} & \textbf{0.60} & \textbf{0.74} \\
mediummean & \textbf{-0.06} & \textbf{0.13} & \textbf{0.13} & \textbf{0.23} & \textbf{0.64} & \textbf{0.90} & \textbf{0.66} & \textbf{0.94} \\
mediumdense & \textbf{-0.07} & \textbf{0.10} & \textbf{0.69} & \textbf{0.80} & \textbf{0.66} & \textbf{0.56} & \textbf{0.69} & \textbf{0.56} \\
hardsparse & \textbf{0.03} & \textbf{0.42} & \textbf{0.10} & \textbf{0.18} & \textbf{0.43} & \textbf{0.72} & \textbf{0.45} & \textbf{0.72} \\
hardmean & \textbf{-0.03} & \textbf{0.11} & \textbf{0.19} & \textbf{0.30} & \textbf{0.23} & \textbf{0.11} & \textbf{0.28} & \textbf{0.25} \\
harddense & \textbf{-0.01} & \textbf{0.14} & \textbf{0.22} & \textbf{0.45} & \textbf{0.32} & \textbf{0.53} & \textbf{0.37} & \textbf{0.67} \\

\midrule
\textbf{MetaDrive Avg} & \textbf{-0.04} & \textbf{0.15} & \textbf{0.31} & \textbf{0.63} & \textbf{0.34} & \textbf{0.48} & \textbf{0.36} & \textbf{0.51} \\

\bottomrule
\end{tabular}
\end{adjustbox}
\end{table*}

\subsubsection{Rollout of shared backbone agents vs. separate ones.}
The results validate our intuition. Using the IQL algorithm, we found that in 23 tasks, the shared IQL reward head incurred lower costs compared to a separate IQL reward agent. Similarly, the shared SAC+BC reward head incurred lower costs in 17 tasks compared to its separate counterpart. Furthermore, the SAC+BC shared reward head collected more rewards in 30 tasks, whereas the separate agent’s disregard for cost information led to sub-optimal performance. This behavior was observed in only 11 tasks when using IQL, highlighting IQL's robustness compared to SAC+BC.

\vspace{1ex}

Additionally, in 23 tasks, the shared IQL cost head collected higher rewards compared to a separate IQL cost agent. Even more impressively, in 34 tasks, the shared SAC+BC cost head collected higher rewards compared to its separate counterpart. These findings underscore the effectiveness of a shared network in enhancing both cost and reward performance, confirming our hypothesis that a shared representation leads to superior outcomes in multi-objective learning scenarios. Results are shown in Tables \ref{tab:shared_sac_cost}, \ref{tab:shared_sac_reward}, \ref{tab:shared_iql_cost}, and \ref{tab:shared_iql_reward}.

\begin{table*}[ht]
    \centering
    \caption{SAC+BC performance comparison of shared backbone cost head and separate cost agent across different environments.}
    \label{tab:shared_sac_cost}
    \begin{adjustbox}{max width=\textwidth}
    \begin{tabular}{l|cccc|cccc}
        \toprule
        & \multicolumn{4}{c|}{Shared Backbone Cost Head} & \multicolumn{4}{c}{Separate Cost Agent} \\
        \cmidrule(lr){2-5} \cmidrule(lr){6-9}
        \textbf{Environment} &  Reward & Normalized Reward & Cost  & Episode Length & Reward Return & Normalized Reward & Cost  & Episode Length \\
        \midrule
        PointButton1 & -0.01 & 0.00 & 18.25 & 1000.00 & -1.57 & -0.04 & 0.05 & 1000.00 \\
        PointButton2 & 5.89 & 0.14 & 29.85 & 1000.00 & -2.32 & -0.05 & 3.70 & 1000.00 \\
        PointCircle1 & 40.49 & 0.49 & 2.90 & 500.00 & 41.63 & 0.52 & 10.40 & 500.00 \\
        PointCircle2 & 38.82 & 0.55 & 9.90 & 500.00 & 35.12 & 0.44 & 0.00 & 500.00 \\
        PointGoal1 & 3.60 & 0.12 & 5.70 & 1000.00 & -3.91 & -0.13 & 0.00 & 1000.00 \\
        PointGoal2 & 4.39 & 0.16 & 13.65 & 1000.00 & -3.64 & -0.13 & 0.45 & 1000.00 \\
        PointPush1 & 2.97 & 0.18 & 17.60 & 1000.00 & -15.80 & -0.96 & 0.00 & 1000.00 \\
        PointPush2 & 1.76 & 0.12 & 24.00 & 1000.00 & -0.50 & -0.03 & 0.90 & 1000.00 \\
        CarButton1 & 0.84 & 0.02 & 72.55 & 1000.00 & -3.06 & -0.07 & 14.95 & 1000.00 \\
        CarButton2 & -9.20 & -0.22 & 23.20 & 1000.00 & -1.25 & -0.03 & 4.45 & 1000.00 \\
        CarCircle1 & 19.30 & 0.67 & 113.20 & 500.00 & 3.67 & -0.26 & 0.00 & 500.00 \\
        CarCircle2 & 19.16 & 0.67 & 176.40 & 500.00 & 9.17 & 0.19 & 0.00 & 500.00 \\
        CarGoal1 & 13.06 & 0.33 & 21.50 & 1000.00 & -3.94 & -0.10 & 0.00 & 1000.00 \\
        CarGoal2 & 7.34 & 0.25 & 39.25 & 1000.00 & -6.46 & -0.22 & 0.00 & 1000.00 \\
        CarPush1 & 3.05 & 0.19 & 19.90 & 1000.00 & -0.74 & -0.05 & 0.00 & 1000.00 \\
        CarPush2 & 1.80 & 0.12 & 23.10 & 1000.00 & -1.21 & -0.08 & 23.15 & 1000.00 \\
        
        SwimmerVelocity & 67.01 & 0.28 & 0.00 & 1000.00 & -2.23 & -0.01 & 0.00 & 1000.00 \\
        HopperVelocity & 641.71 & 0.32 & 10.40 & 387.00 & 29.29 & 0.00 & 0.00 & 25.15 \\
        HalfCheetahVelocity & 2397.03 & 0.85 & 0.00 & 1000.00 & -260.81 & -0.10 & 0.00 & 1000.00 \\
        Walker2dVelocity & 2688.10 & 0.79 & 0.00 & 1000.00 & -5.78 & -0.01 & 0.00 & 7.00 \\
        AntVelocity & 2047.74 & 0.69 & 0.00 & 1000.00 & -1902.42 & -0.64 & 0.00 & 1000.00 \\
        \midrule
        BallRun & 140.58 & 0.09 & 0.00 & 100.00 & 387.43 & 0.28 & 0.00 & 100.00 \\
        CarRun & 486.34 & 0.76 & 0.00 & 200.00 & 286.02 & 0.22 & 0.00 & 200.00 \\
        DroneRun & 365.30 & 0.53 & 0.00 & 200.00 & 41.21 & 0.05 & 0.00 & 200.00 \\
        AntRun & 425.31 & 0.45 & 0.05 & 200.00 & 31.50 & 0.03 & 0.00 & 200.00 \\
        BallCircle & 475.28 & 0.54 & 2.30 & 200.00 & 0.43 & 0.00 & 0.00 & 200.00 \\
        CarCircle & 279.19 & 0.52 & 0.00 & 300.00 & 166.77 & 0.31 & 0.00 & 300.00 \\
        DroneCircle & 481.65 & 0.35 & 0.00 & 300.00 & -23.62 & -0.29 & 36.95 & 147.40 \\
        AntCircle & 207.09 & 0.45 & 1.20 & 477.55 & 0.59 & 0.00 & 0.00 & 156.05 \\
        
        \midrule
        EasySparse & 374.23 & 0.87 & 53.13 & 815.75 & -4.16 & -0.06 & 0.00 & 1000.00 \\
        EasyMean & 99.72 & 0.19 & 2.24 & 605.95 & -4.30 & -0.07 & 1.00 & 18.00 \\
        EasyDense & 10.01 & -0.02 & 1.80 & 73.50 & 12.04 & -0.02 & 1.00 & 85.75 \\
        MediumSparse & 240.24 & 0.88 & 19.44 & 724.70 & -3.48 & -0.08 & 1.00 & 21.00 \\
        MediumMean & 10.89 & -0.02 & 1.00 & 119.80 & 0.49 & -0.07 & 0.00 & 1000.00 \\
        MediumDense & 182.84 & 0.66 & 6.83 & 717.25 & -3.95 & -0.07 & 1.00 & 134.00 \\
        HardSparse & 9.80 & -0.02 & 1.00 & 133.55 & 9.01 & -0.02 & 1.00 & 92.00 \\
        HardMean & 64.68 & 0.10 & 2.52 & 532.95 & -0.13 & -0.04 & 1.00 & 108.00 \\
        HardDense & 109.53 & 0.20 & 3.57 & 837.05 & -6.21 & -0.05 & 1.00 & 311.00 \\
        \bottomrule
    \end{tabular}
    
\end{adjustbox}
\end{table*}

\begin{table*}[ht]
    \centering
    \caption{SAC+BC performance comparison of shared backbone reward head and separate reward agent across different environments.}
    \label{tab:shared_sac_reward}
    \begin{adjustbox}{max width=\textwidth}
    \begin{tabular}{l|cccc|cccc}
        \toprule
        & \multicolumn{4}{c|}{Shared Backbone Reward Head} & \multicolumn{4}{c}{Separate Reward Agent} \\
        \cmidrule(lr){2-5} \cmidrule(lr){6-9}
        \textbf{Environment} &  Reward & Normalized Reward & Cost  & Episode Length & Reward Return & Normalized Reward & Cost  & Episode Length \\
        \midrule
        PointButton1 & 3.43 & 0.08 & 16.35 & 1000.00 & 3.28 & 0.08 & 26.65 & 1000.00 \\
        PointButton2 & 14.56 & 0.34 & 77.10 & 1000.00 & 0.59 & 0.01 & 107.50 & 1000.00 \\
        PointCircle1 & 47.99 & 0.67 & 189.50 & 500.00 & 49.60 & 0.71 & 193.75 & 500.00 \\
        PointCircle2 & 47.17 & 0.80 & 212.45 & 500.00 & 41.69 & 0.63 & 171.60 & 500.00 \\
        PointGoal1 & 20.11 & 0.67 & 31.45 & 1000.00 & 19.13 & 0.64 & 49.90 & 1000.00 \\
        PointGoal2 & 15.26 & 0.55 & 67.30 & 1000.00 & -1.52 & -0.05 & 72.65 & 1000.00 \\
        PointPush1 & 2.90 & 0.18 & 22.50 & 1000.00 & 0.32 & 0.02 & 98.40 & 1000.00 \\
        PointPush2 & 2.49 & 0.17 & 29.85 & 1000.00 & -1.51 & -0.10 & 28.75 & 1000.00 \\
        
        CarButton1 & 0.97 & 0.02 & 47.75 & 1000.00 & 4.43 & 0.10 & 276.95 & 1000.00 \\
        CarButton2 & -4.37 & -0.10 & 52.80 & 1000.00 & 4.38 & 0.10 & 297.45 & 1000.00 \\
        CarCircle1 & 19.94 & 0.70 & 207.10 & 500.00 & 14.15 & 0.36 & 51.95 & 500.00 \\
        CarCircle2 & 20.88 & 0.76 & 275.55 & 500.00 & 15.89 & 0.52 & 244.90 & 500.00 \\
        CarGoal1 & 15.67 & 0.39 & 35.65 & 1000.00 & 2.23 & 0.06 & 89.15 & 1000.00 \\
        CarGoal2 & 6.46 & 0.22 & 28.60 & 1000.00 & -1.34 & -0.05 & 69.85 & 1000.00 \\
        CarPush1 & 2.94 & 0.18 & 16.90 & 1000.00 & 0.50 & 0.03 & 0.00 & 1000.00 \\
        CarPush2 & 2.04 & 0.13 & 26.15 & 1000.00 & 0.22 & 0.01 & 139.55 & 1000.00 \\
        
        SwimmerVelocity & 140.33 & 0.59 & 177.35 & 1000.00 & 72.80 & 0.30 & 95.45 & 1000.00 \\
        HopperVelocity & 1753.94 & 0.92 & 530.80 & 1000.00 & 752.65 & 0.38 & 189.20 & 198.70 \\
        HalfCheetahVelocity & 2744.72 & 0.98 & 86.10 & 1000.00 & 4488.50 & 1.60 & 975.70 & 1000.00 \\
        Walker2dVelocity & 2697.91 & 0.79 & 0.00 & 1000.00 & 181.58 & 0.05 & 14.60 & 88.65 \\
        AntVelocity & 2887.08 & 0.97 & 30.90 & 1000.00 & -2319.29 & -0.78 & 0.00 & 1000.00 \\
        \midrule
        BallCircle & 786.73 & 0.89 & 63.95 & 200.00 & 765.98 & 0.87 & 60.85 & 200.00 \\
        CarCircle & 430.84 & 0.81 & 93.65 & 300.00 & 455.02 & 0.85 & 90.15 & 300.00 \\
        DroneCircle & 851.22 & 0.82 & 90.50 & 300.00 & 48.63 & -0.20 & 13.35 & 65.40 \\
        AntCircle & 376.99 & 0.82 & 139.65 & 462.05 & 15.13 & 0.03 & 14.10 & 307.70 \\
        BallRun & 1419.25 & 1.07 & 86.50 & 100.00 & 1459.21 & 1.10 & 92.00 & 100.00 \\
        CarRun & 567.43 & 0.98 & 0.50 & 200.00 & 576.56 & 1.01 & 79.35 & 200.00 \\
        DroneRun & 487.25 & 0.71 & 100.60 & 200.00 & 570.95 & 0.83 & 145.40 & 189.75 \\
        AntRun & 667.27 & 0.70 & 74.55 & 200.00 & 42.55 & 0.04 & 0.00 & 200.00 \\
        \midrule
        EasySparse & 134.53 & 0.28 & 28.91 & 254.85 & -4.17 & -0.06 & 1.00 & 16.00 \\
        EasyMean & 59.87 & 0.09 & 11.39 & 152.20 & 6.04 & -0.04 & 3.25 & 39.00 \\
        EasyDense & 59.50 & 0.10 & 8.94 & 228.05 & -4.14 & -0.06 & 1.00 & 16.00 \\
        MediumSparse & 108.45 & 0.36 & 25.27 & 160.40 & 9.53 & -0.03 & 4.02 & 42.00 \\
        MediumMean & 8.46 & -0.03 & 2.84 & 46.00 & -2.64 & -0.08 & 1.09 & 21.00 \\
        MediumDense & 15.33 & 0.00 & 3.43 & 74.00 & 3.71 & -0.04 & 2.65 & 37.00 \\
        HardSparse & 7.15 & -0.02 & 1.00 & 110.00 & 1.60 & -0.03 & 2.10 & 35.00 \\
        HardMean & 15.38 & -0.01 & 3.37 & 67.00 & -0.91 & -0.04 & 1.35 & 25.00 \\
        HardDense & 82.38 & 0.14 & 15.24 & 159.65 & -2.65 & -0.04 & 1.09 & 21.00 \\
        \bottomrule
    \end{tabular}
    
\end{adjustbox}
\end{table*}

\begin{table*}[ht]
    \centering
    \caption{IQL performance comparison of shared backbone cost head and separate cost agent across different environments.}
    \label{tab:shared_iql_cost}
    \begin{adjustbox}{max width=\textwidth}
    \begin{tabular}{l|cccc|cccc}
        \toprule
        & \multicolumn{4}{c|}{Shared Backbone Cost Head} & \multicolumn{4}{c}{Separate Cost Agent} \\
        \cmidrule(lr){2-5} \cmidrule(lr){6-9}
        \textbf{Environment} &  Reward & Normalized Reward & Cost  & Episode Length & Reward Return & Normalized Reward & Cost  & Episode Length \\
        \midrule
        PointButton1 & 1.44 & 0.03 & 21.22 & 1000.00 & -0.75 & -0.02 & 10.43 & 1000.00 \\
        PointButton2 & 3.07 & 0.07 & 21.00 & 1000.00 & 4.53 & 0.11 & 17.73 & 1000.00 \\
        PointCircle1 & 24.33 & 0.10 & 11.37 & 500.00 & 28.70 & 0.21 & 15.60 & 500.00 \\
        PointCircle2 & 34.56 & 0.42 & 0.88 & 500.00 & 32.91 & 0.37 & 0.00 & 500.00 \\
        PointGoal1 & 5.90 & 0.20 & 9.18 & 1000.00 & 6.94 & 0.23 & 8.33 & 1000.00 \\
        PointGoal2 & 4.32 & 0.16 & 10.03 & 1000.00 & 4.45 & 0.16 & 12.78 & 1000.00 \\
        PointPush1 & 2.33 & 0.14 & 8.52 & 1000.00 & 1.34 & 0.08 & 14.20 & 1000.00 \\
        PointPush2 & 1.79 & 0.12 & 21.45 & 1000.00 & 1.34 & 0.09 & 12.85 & 1000.00 \\
        
        CarButton1 & -0.57 & -0.01 & 8.07 & 1000.00 & -1.22 & -0.03 & 12.62 & 1000.00 \\
        CarButton2 & -3.28 & -0.08 & 6.78 & 1000.00 & -5.19 & -0.12 & 15.62 & 1000.00 \\
        CarCircle1 & 15.69 & 0.45 & 34.85 & 500.00 & 15.41 & 0.44 & 37.63 & 500.00 \\
        CarCircle2 & 14.23 & 0.44 & 59.22 & 500.00 & 13.62 & 0.41 & 38.67 & 500.00 \\
        CarGoal1 & 14.21 & 0.36 & 14.68 & 1000.00 & 11.84 & 0.30 & 9.35 & 1000.00 \\
        CarGoal2 & 3.45 & 0.12 & 14.83 & 1000.00 & 2.11 & 0.07 & 7.18 & 1000.00 \\
        CarPush1 & 2.85 & 0.17 & 7.53 & 1000.00 & 2.21 & 0.13 & 5.90 & 1000.00 \\
        CarPush2 & 0.77 & 0.05 & 16.13 & 1000.00 & 0.62 & 0.04 & 15.00 & 1000.00 \\

        SwimmerVelocity & 96.85 & 0.41 & 28.60 & 1000.00 & 78.59 & 0.33 & 75.28 & 1000.00 \\
        HopperVelocity & 486.36 & 0.24 & 8.60 & 348.05 & 698.29 & 0.35 & 1.20 & 592.55 \\
        HalfCheetahVelocity & 2413.98 & 0.86 & 0.42 & 1000.00 & 2378.76 & 0.85 & 0.12 & 1000.00 \\
        Walker2dVelocity & 2674.55 & 0.78 & 0.07 & 1000.00 & 2682.83 & 0.78 & 2.97 & 1000.00 \\
        AntVelocity & 2499.16 & 0.84 & 0.77 & 1000.00 & 2504.33 & 0.84 & 0.20 & 1000.00 \\

        \midrule
        BallCircle & 223.10 & 0.25 & 0.50 & 200.00 & 135.04 & 0.15 & 5.07 & 200.00 \\
        CarCircle & 50.83 & 0.09 & 0.00 & 300.00 & 27.68 & 0.05 & 0.00 & 300.00 \\
        DroneCircle & 366.74 & 0.20 & 0.00 & 300.00 & 364.69 & 0.20 & 0.23 & 298.15 \\
        AntCircle & 131.15 & 0.28 & 0.00 & 444.50 & 125.54 & 0.27 & 0.08 & 438.78 \\
        BallRun & 110.83 & 0.06 & 0.00 & 100.00 & 169.66 & 0.11 & 0.00 & 100.00 \\
        CarRun & 479.18 & 0.74 & 0.60 & 200.00 & 512.50 & 0.83 & 0.00 & 200.00 \\
        DroneRun & 258.95 & 0.37 & 71.87 & 199.18 & 316.17 & 0.45 & 38.65 & 199.70 \\
        AntRun & 439.12 & 0.46 & 6.05 & 200.00 & 523.04 & 0.55 & 10.28 & 200.00 \\
        
        \midrule
        EasySparse & 46.93 & 0.07 & 3.20 & 376.07 & 115.36 & 0.23 & 3.17 & 782.12 \\
        EasyMean & 23.90 & 0.00 & 3.66 & 123.93 & 35.04 & 0.03 & 3.14 & 172.10 \\
        EasyDense & 61.58 & 0.10 & 1.69 & 452.87 & 98.31 & 0.19 & 0.84 & 794.32 \\
        MediumSparse & 118.93 & 0.40 & 5.29 & 659.57 & 167.06 & 0.59 & 16.50 & 514.97 \\
        MediumMean & 137.99 & 0.48 & 12.51 & 727.23 & 117.97 & 0.40 & 9.86 & 627.22 \\
        MediumDense & 179.50 & 0.65 & 6.84 & 824.30 & 95.94 & 0.32 & 1.78 & 593.70 \\
        HardSparse & 193.47 & 0.38 & 3.84 & 897.07 & 49.76 & 0.07 & 5.48 & 168.00 \\
        HardMean & 107.38 & 0.19 & 0.54 & 928.90 & 82.52 & 0.14 & 4.73 & 373.23 \\
        HardDense & 148.81 & 0.28 & 4.88 & 911.10 & 105.56 & 0.19 & 15.80 & 252.83 \\
        \bottomrule
    \end{tabular}
    
\end{adjustbox}
\end{table*}

\begin{table*}[ht]
    \centering
    \caption{IQL performance comparison of shared backbone reward head and separate reward agent across different environments.}
    \label{tab:shared_iql_reward}
    \begin{adjustbox}{max width=\textwidth}
    \begin{tabular}{l|cccc|cccc}
        \toprule
        & \multicolumn{4}{c|}{Shared Backbone Reward Head} & \multicolumn{4}{c}{Separate Reward Agent} \\
        \cmidrule(lr){2-5} \cmidrule(lr){6-9}
        \textbf{Environment} &  Reward & Normalized Reward & Cost  & Episode Length & Reward Return & Normalized Reward & Cost  & Episode Length \\
        \midrule
        PointButton1 & 4.79 & 0.12 & 30.62 & 1000.00 & 5.18 & 0.13 & 33.47 & 1000.00 \\
        PointButton2 & 11.95 & 0.28 & 65.03 & 1000.00 & 12.32 & 0.29 & 69.48 & 1000.00 \\
        PointCircle1 & 50.38 & 0.73 & 176.30 & 500.00 & 51.06 & 0.74 & 192.32 & 500.00 \\
        PointCircle2 & 47.12 & 0.79 & 232.92 & 500.00 & 45.49 & 0.75 & 199.08 & 500.00 \\
        PointGoal1 & 20.28 & 0.67 & 38.15 & 1000.00 & 17.18 & 0.57 & 27.38 & 1000.00 \\
        PointGoal2 & 14.44 & 0.52 & 88.05 & 1000.00 & 13.70 & 0.49 & 73.12 & 1000.00 \\
        PointPush1 & 3.25 & 0.20 & 29.98 & 1000.00 & 3.01 & 0.18 & 30.65 & 1000.00 \\
        PointPush2 & 1.75 & 0.12 & 29.38 & 1000.00 & 1.73 & 0.12 & 40.10 & 1000.00 \\
        
        CarButton1 & 0.24 & 0.01 & 54.80 & 1000.00 & 2.97 & 0.07 & 43.73 & 1000.00 \\
        CarButton2 & -3.71 & -0.09 & 40.52 & 1000.00 & -0.83 & -0.02 & 58.40 & 1000.00 \\
        CarCircle1 & 20.38 & 0.73 & 190.42 & 500.00 & 20.47 & 0.74 & 197.78 & 500.00 \\
        CarCircle2 & 20.28 & 0.73 & 258.28 & 500.00 & 20.98 & 0.76 & 270.90 & 500.00 \\
        CarGoal1 & 16.47 & 0.41 & 19.22 & 1000.00 & 13.68 & 0.34 & 26.05 & 1000.00 \\
        CarGoal2 & 8.12 & 0.28 & 37.63 & 1000.00 & 6.29 & 0.22 & 35.67 & 1000.00 \\
        CarPush1 & 2.90 & 0.18 & 21.03 & 1000.00 & 3.50 & 0.21 & 17.03 & 1000.00 \\
        CarPush2 & 1.05 & 0.07 & 27.12 & 1000.00 & 1.78 & 0.12 & 28.52 & 1000.00 \\

        SwimmerVelocity & 53.22 & 0.22 & 72.37 & 1000.00 & 48.98 & 0.20 & 56.58 & 1000.00 \\        
        HopperVelocity & 654.39 & 0.33 & 158.12 & 191.83 & 673.98 & 0.34 & 145.85 & 195.38 \\
        HalfCheetahVelocity & 2793.41 & 1.00 & 250.60 & 1000.00 & 2797.74 & 1.00 & 187.90 & 1000.00 \\
        Walker2dVelocity & 3078.16 & 0.90 & 206.02 & 959.28 & 3122.61 & 0.91 & 196.93 & 988.68 \\
        AntVelocity & 2895.56 & 0.97 & 140.27 & 1000.00 & 2964.54 & 1.00 & 208.55 & 1000.00 \\
        \midrule
        BallCircle & 828.74 & 0.94 & 59.08 & 200.00 & 838.51 & 0.95 & 60.42 & 200.00 \\
        CarCircle & 513.94 & 0.96 & 92.77 & 300.00 & 508.34 & 0.95 & 91.67 & 300.00 \\
        AntCircle & 309.75 & 0.67 & 95.95 & 451.78 & 357.56 & 0.78 & 124.58 & 486.78 \\
        DroneCircle & 906.13 & 0.89 & 99.17 & 300.00 & 952.44 & 0.94 & 98.25 & 300.00 \\
        BallRun & 1546.65 & 1.17 & 92.33 & 100.00 & 1568.23 & 1.19 & 92.33 & 100.00 \\
        CarRun & 564.23 & 0.97 & 5.12 & 200.00 & 564.44 & 0.97 & 9.20 & 200.00 \\
        DroneRun & 391.63 & 0.57 & 31.57 & 200.00 & 422.06 & 0.61 & 22.83 & 200.00 \\
        AntRun & 925.83 & 0.97 & 146.22 & 199.58 & 961.77 & 1.01 & 159.13 & 200.00 \\
        
        \midrule
        EasySparse & 119.32 & 0.24 & 25.35 & 230.28 & 408.73 & 0.96 & 69.40 & 744.73 \\
        EasyMean & 40.32 & 0.05 & 10.02 & 84.27 & 303.79 & 0.70 & 52.96 & 728.45 \\
        EasyDense & 88.30 & 0.17 & 11.30 & 348.38 & 257.20 & 0.58 & 41.75 & 750.50 \\
        MediumSparse & 136.17 & 0.47 & 15.16 & 393.50 & 246.68 & 0.91 & 35.13 & 574.83 \\
        MediumMean & 213.06 & 0.78 & 33.24 & 490.97 & 196.74 & 0.71 & 28.82 & 479.43 \\
        MediumDense & 215.81 & 0.79 & 26.60 & 583.42 & 195.87 & 0.71 & 32.88 & 352.27 \\
        HardSparse & 225.52 & 0.44 & 29.73 & 576.53 & 231.79 & 0.46 & 30.85 & 667.65 \\
        HardMean & 158.70 & 0.30 & 11.33 & 715.70 & 208.21 & 0.40 & 25.68 & 605.25 \\
        HardDense & 176.90 & 0.34 & 19.76 & 554.32 & 221.44 & 0.44 & 31.67 & 561.67 \\
        
        \bottomrule
    \end{tabular}
    
\end{adjustbox}
\end{table*}

\subsection{Cost Limit Ablations} \label{ap:cost_lim_ablations}

\vspace{1ex}

Table \ref{tab:low_costs} compares the performance of CAPS(IQL) and CDT across different extra cost configurations \{5, 10\}, \{15, 30\}, and \{30, 60\}, revealing that CAPS(IQL) consistently outperforms CDT in maintaining safety while delivering competitive rewards. CAPS(IQL) effectively manages costs across all configurations, ensuring they remain within the safety threshold, even as costs decreases. In contrast, CDT, while occasionally achieving higher rewards, often does so at the expense of significantly higher costs, leading to many unsafe outcomes. Specifically, CAPS(IQL) achieves safe outcomes in 18, 31, and 33 instances for the three different cost settings, respectively, while CDT manages safety only 11, 19, and 22 times under the same conditions. This demonstrates that CAPS(IQL) is a more reliable and robust method for balancing performance with safety, making it particularly suitable for environments where cost management and risk mitigation are critical. Table \ref{tab:sac_low_costs} presents the SAC+BC instantiation for the extra cost configurations.

Note that CDT requires specifying return and cost targets to generate trajectories. However, the authors do not provide a clear methodology for selecting these return targets; the values are hand-coded for each environment without any explanation of the rationale behind these choices. To address this, we used the provided return targets for the cost sets \{10, 20, 40\} and \{20, 40, 80\} to interpolate or extrapolate return targets for the additional cost sets \{5, 15, 30\} and \{10, 30, 60\}. This approach helps ensure consistency across different cost sets, although it may introduce some uncertainty in the resulting trajectories.

\begin{table*}[ht!]
\centering
\caption{Performance across different extra cost configurations for CAPS(IQL) and CDT. The cost threshold is 1. The $\uparrow$ symbol denotes that a higher reward is better. The $\downarrow$ symbol denotes that a lower cost (up to threshold 1) is better. Each value is averaged over 20 evaluation episodes, and 3 random seeds. {\bf Bold}: Safe agents whose normalized cost $\leq$ 1. \textcolor{gray}{Gray}: Unsafe agents.}
\label{tab:low_costs}
\begin{adjustbox}{max width=\textwidth}
\begin{tabular}{l|cc|cc|cc|cc|cc|cc}
\toprule
\textbf{Cost Configuration} & \multicolumn{4}{c|}{\{5, 10\}} & \multicolumn{4}{c|}{\{15, 30\}} & \multicolumn{4}{c}{\{30, 60\}} \\
\cmidrule(lr){2-5} \cmidrule(lr){6-9} \cmidrule(lr){10-13} 
& \multicolumn{2}{c|}{\textbf{CAPS IQL}} & \multicolumn{2}{c|}{\textbf{CDT}} & \multicolumn{2}{c|}{\textbf{CAPS IQL}} & \multicolumn{2}{c|}{\textbf{CDT}} & \multicolumn{2}{c|}{\textbf{CAPS IQL}} & \multicolumn{2}{c}{\textbf{CDT}} \\
\textbf{Environment} & reward $\uparrow$ & cost $\downarrow$ & reward $\uparrow$ & cost $\downarrow$ & reward $\uparrow$ & cost $\downarrow$ & reward $\uparrow$ & cost $\downarrow$ & reward $\uparrow$ & cost $\downarrow$ & reward $\uparrow$ & cost $\downarrow$  \\
\midrule
PointButton1  & \textcolor{gray}{0.02} & \textcolor{gray}{1.36} & \textcolor{gray}{0.51} & \textcolor{gray}{9.56} & \textbf{0.02} & \textbf{0.35} & \textcolor{gray}{0.53} & \textcolor{gray}{3.60} & \textbf{0.07} & \textbf{0.33} & \textcolor{gray}{0.54} & \textcolor{gray}{1.93} \\
PointButton2  & \textcolor{gray}{0.14} & \textcolor{gray}{2.84} & \textcolor{gray}{0.42} & \textcolor{gray}{9.33} & \textcolor{gray}{0.15} & \textcolor{gray}{1.09} & \textcolor{gray}{0.47} & \textcolor{gray}{2.95} & \textbf{0.13} & \textbf{0.52} & \textcolor{gray}{0.43} & \textcolor{gray}{1.62} \\
PointCircle1  & \textcolor{gray}{0.27} & \textcolor{gray}{1.27} & \textbf{0.54} & \textbf{0.58} & \textbf{0.49} & \textbf{0.73} & \textbf{0.56} & \textbf{0.77} & \textbf{0.53} & \textbf{0.64} & \textbf{0.59} & \textbf{0.92} \\
PointCircle2  & \textbf{0.42} & \textbf{0.11} & \textcolor{gray}{0.60} & \textcolor{gray}{1.93} & \textbf{0.44} & \textbf{0.58} & \textcolor{gray}{0.62} & \textcolor{gray}{1.46} & \textcolor{gray}{0.59} & \textcolor{gray}{1.01} & \textcolor{gray}{0.64} & \textcolor{gray}{1.16} \\
PointGoal1  & \textcolor{gray}{0.32} & \textcolor{gray}{1.54} & \textcolor{gray}{0.69} & \textcolor{gray}{3.17} & \textbf{0.46} & \textbf{0.75} & \textcolor{gray}{0.69} & \textcolor{gray}{1.30} & \textbf{0.57} & \textbf{0.50} & \textbf{0.73} & \textbf{0.62} \\
PointGoal2  & \textcolor{gray}{0.22} & \textcolor{gray}{1.12} & \textcolor{gray}{0.43} & \textcolor{gray}{5.30} & \textbf{0.27} & \textbf{0.72} & \textcolor{gray}{0.49} & \textcolor{gray}{2.12} & \textbf{0.37} & \textbf{0.62} & \textcolor{gray}{0.59} & \textcolor{gray}{1.39} \\
PointPush1  & \textbf{0.12} & \textbf{0.52} & \textcolor{gray}{0.23} & \textcolor{gray}{3.02} & \textbf{0.16} & \textbf{0.54} & \textcolor{gray}{0.25} & \textcolor{gray}{1.19} & \textbf{0.2} & \textbf{0.35} & \textbf{0.21} & \textbf{0.45} \\
PointPush2  & \textcolor{gray}{0.09} & \textcolor{gray}{2.08} & \textcolor{gray}{0.20} & \textcolor{gray}{4.54} & \textbf{0.10} & \textbf{0.90} & \textcolor{gray}{0.20} & \textcolor{gray}{1.45} & \textbf{0.15} & \textbf{0.45} & \textcolor{gray}{0.23} & \textcolor{gray}{1.02} \\

CarButton1  & \textbf{-0.04} & \textbf{0.62} & \textcolor{gray}{0.20} & \textcolor{gray}{9.04} & \textbf{-0.02} & \textbf{0.24} & \textcolor{gray}{0.20} & \textcolor{gray}{3.32} & \textbf{0.01} & \textbf{0.24} & \textcolor{gray}{0.18} & \textcolor{gray}{1.67} \\
CarButton2  & \textcolor{gray}{-0.12} & \textcolor{gray}{1.13} & \textcolor{gray}{0.25} & \textcolor{gray}{14.47} & \textbf{-0.11} & \textbf{0.36} & \textcolor{gray}{0.24} & \textcolor{gray}{4.93} & \textbf{-0.08} & \textbf{0.20} & \textcolor{gray}{0.13} & \textcolor{gray}{1.95} \\
CarCircle1  & \textcolor{gray}{0.46} & \textcolor{gray}{4.21} & \textcolor{gray}{0.54} & \textcolor{gray}{8.50} & \textcolor{gray}{0.49} & \textcolor{gray}{1.55} & \textcolor{gray}{0.54} & \textcolor{gray}{3.19} & \textcolor{gray}{0.59} & \textcolor{gray}{1.30} & \textcolor{gray}{0.65} & \textcolor{gray}{2.33} \\
CarCircle2  & \textcolor{gray}{0.45} & \textcolor{gray}{4.94} & \textcolor{gray}{0.63} & \textcolor{gray}{13.20} & \textcolor{gray}{0.47} & \textcolor{gray}{1.53} & \textcolor{gray}{0.65} & \textcolor{gray}{4.89} & \textcolor{gray}{0.54} & \textcolor{gray}{1.21} & \textcolor{gray}{0.69} & \textcolor{gray}{3.11} \\
CarGoal1  & \textcolor{gray}{0.29} & \textcolor{gray}{1.24} & \textcolor{gray}{0.62} & \textcolor{gray}{3.27} & \textbf{0.32} & \textbf{0.47} & \textcolor{gray}{0.66} & \textcolor{gray}{1.27} & \textbf{0.31} & \textbf{0.27} & \textbf{0.67} & \textbf{0.75} \\
CarGoal2  & \textcolor{gray}{0.12} & \textcolor{gray}{2.52} & \textcolor{gray}{0.43} & \textcolor{gray}{7.01} & \textbf{0.13} & \textbf{0.75} & \textcolor{gray}{0.44} & \textcolor{gray}{2.20} & \textbf{0.19} & \textbf{0.45} & \textcolor{gray}{0.57} & \textcolor{gray}{1.43} \\
CarPush1  & \textbf{0.17} & \textbf{0.48} & \textcolor{gray}{0.28} & \textcolor{gray}{2.23} & \textbf{0.19} & \textbf{0.25} & \textbf{0.29} & \textbf{0.78} & \textbf{0.19} & \textbf{0.13} & \textbf{0.33} & \textbf{0.54} \\
CarPush2  & \textcolor{gray}{0.07} & \textcolor{gray}{2.21} & \textcolor{gray}{0.15} & \textcolor{gray}{6.10} & \textbf{0.10} & \textbf{0.85} & \textcolor{gray}{0.15} & \textcolor{gray}{2.11} & \textbf{0.06} & \textbf{0.37} & \textcolor{gray}{0.20} & \textcolor{gray}{1.21} \\

SwimmerVelocity  & \textcolor{gray}{0.45} & \textcolor{gray}{3.35} & \textbf{0.66} & \textbf{0.98} & \textcolor{gray}{0.49} & \textcolor{gray}{2.47} & \textbf{0.68} & \textbf{0.96} & \textcolor{gray}{0.44} & \textcolor{gray}{1.47} & \textbf{0.69} & \textbf{0.95} \\
HopperVelocity  & \textcolor{gray}{0.26} & \textcolor{gray}{1.40} & \textcolor{gray}{0.69} & \textcolor{gray}{1.16} & \textbf{0.34} & \textbf{0.70} & \textbf{0.75} & \textbf{0.79} & \textbf{0.51} & \textbf{0.68} & \textbf{0.71} & \textbf{0.73} \\
HalfCheetahVelocity  & \textbf{0.87} & \textbf{0.31} & \textbf{0.98} & \textbf{0.68} & \textbf{0.92} & \textbf{0.77} & \textbf{0.98} & \textbf{0.18} & \textbf{0.97} & \textbf{0.84} & \textbf{0.98} & \textbf{0.11} \\
Walker2dVelocity  & \textbf{0.79} & \textbf{0.09} & \textbf{0.80} & \textbf{0.16} & \textbf{0.8} & \textbf{0.54} & \textbf{0.78} & \textbf{0.15} & \textbf{0.81} & \textbf{0.86} & \textbf{0.78} & \textbf{0.14} \\
AntVelocity  & \textbf{0.87} & \textbf{0.22} & \textbf{0.98} & \textbf{0.71} & \textbf{0.94} & \textbf{0.60} & \textbf{0.99} & \textbf{0.52} & \textbf{0.99} & \textbf{0.72} & \textbf{0.99} & \textbf{0.46} \\

\midrule
BallCircle  & \textbf{0.31} & \textbf{0.00} & \textcolor{gray}{0.70} & \textcolor{gray}{2.25} & \textbf{0.65} & \textbf{0.53} & \textcolor{gray}{0.75} & \textcolor{gray}{1.28} & \textbf{0.76} & \textbf{0.77} & \textbf{0.81} & \textbf{0.97} \\
CarCircle  & \textbf{0.41} & \textbf{0.02} & \textcolor{gray}{0.72} & \textcolor{gray}{2.01} & \textbf{0.7} & \textbf{0.62} & \textbf{0.72} & \textbf{0.87} & \textbf{0.72} & \textbf{0.81} & \textbf{0.77} & \textbf{0.96} \\
DroneCircle  & \textbf{0.33} & \textbf{0.00} & \textcolor{gray}{0.55} & \textcolor{gray}{1.39} & \textbf{0.54} & \textbf{0.64} & \textcolor{gray}{0.59} & \textcolor{gray}{1.14} & \textbf{0.58} & \textbf{0.81} & \textcolor{gray}{0.64} & \textcolor{gray}{1.07} \\
AntCircle  & \textbf{0.31} & \textbf{0.11} & \textcolor{gray}{0.43} & \textcolor{gray}{6.21} & \textbf{0.34} & \textbf{0.08} & \textcolor{gray}{0.45} & \textcolor{gray}{2.76} & \textbf{0.45} & \textbf{0.20} & \textcolor{gray}{0.52} & \textcolor{gray}{1.60} \\
BallRun  & \textbf{0.07} & \textbf{0.00} & \textcolor{gray}{0.32} & \textcolor{gray}{2.26} & \textcolor{gray}{0.11} & \textcolor{gray}{1.52} & \textbf{0.32} & \textbf{0.84} & \textcolor{gray}{0.22} & \textcolor{gray}{1.19} & \textbf{0.38} & \textbf{0.95} \\
CarRun  & \textbf{0.97} & \textbf{0.41} & \textbf{0.99} & \textbf{0.94} & \textbf{0.97} & \textbf{0.26} & \textbf{0.99} & \textbf{0.66} & \textbf{0.98} & \textbf{0.23} & \textbf{0.99} & \textbf{0.63} \\
DroneRun  & \textcolor{gray}{0.4} & \textcolor{gray}{12.06} & \textbf{0.51} & \textbf{0.54} & \textcolor{gray}{0.41} & \textcolor{gray}{3.99} & \textbf{0.59} & \textbf{0.49} & \textbf{0.51} & \textbf{0.45} & \textcolor{gray}{0.62} & \textcolor{gray}{1.32} \\
AntRun  & \textcolor{gray}{0.47} & \textcolor{gray}{1.39} & \textcolor{gray}{0.70} & \textcolor{gray}{1.69} & \textbf{0.59} & \textbf{0.90} & \textbf{0.72} & \textbf{0.85} & \textbf{0.69} & \textbf{0.90} & \textbf{0.73} & \textbf{0.79} \\

\midrule
EasySparse  & \textbf{0.09} & \textbf{0.77} & \textbf{0.07} & \textbf{0.00} & \textbf{0.10} & \textbf{0.32} & \textbf{0.18} & \textbf{0.00} & \textbf{0.19} & \textbf{0.54} & \textbf{0.52} & \textbf{0.74} \\
EasyMean  & \textbf{0.00} & \textbf{0.67} & \textcolor{gray}{0.40} & \textcolor{gray}{3.69} & \textbf{0.00} & \textbf{0.22} & \textcolor{gray}{0.45} & \textcolor{gray}{1.31} & \textbf{0.01} & \textbf{0.11} & \textbf{0.49} & \textbf{0.71} \\
EasyDense  & \textbf{0.10} & \textbf{0.53} & \textbf{0.29} & \textbf{0.21} & \textbf{0.11} & \textbf{0.18} & \textbf{0.42} & \textbf{0.25} & \textbf{0.10} & \textbf{0.12} & \textcolor{gray}{0.44} & \textcolor{gray}{1.19} \\
MediumSparse  & \textcolor{gray}{0.51} & \textcolor{gray}{1.95} & \textcolor{gray}{0.44} & \textcolor{gray}{1.16} & \textbf{0.54} & \textbf{0.78} & \textcolor{gray}{0.54} & \textcolor{gray}{1.22} & \textbf{0.63} & \textbf{0.50} & \textcolor{gray}{0.65} & \textcolor{gray}{1.05} \\
MediumMean  & \textcolor{gray}{0.53} & \textcolor{gray}{2.55} & \textcolor{gray}{0.55} & \textcolor{gray}{1.57} & \textcolor{gray}{0.64} & \textcolor{gray}{1.04} & \textbf{0.35} & \textbf{0.99} & \textbf{0.73} & \textbf{0.64} & \textbf{0.36} & \textbf{0.61} \\
MediumDense  & \textcolor{gray}{0.66} & \textcolor{gray}{1.66} & \textbf{0.12} & \textbf{0.17} & \textbf{0.62} & \textbf{0.59} & \textbf{0.15} & \textbf{0.08} & \textbf{0.75} & \textbf{0.51} & \textbf{0.31} & \textbf{0.49} \\
hardsparse  & \textcolor{gray}{0.41} & \textcolor{gray}{2.15} & \textcolor{gray}{0.24} & \textcolor{gray}{1.85} & \textbf{0.45} & \textbf{0.86} & \textbf{0.20} & \textbf{0.20} & \textbf{0.48} & \textbf{0.53} & \textbf{0.30} & \textbf{0.89} \\
hardmean  & \textbf{0.20} & \textbf{0.33} & \textbf{0.07} & \textbf{0.07} & \textbf{0.25} & \textbf{0.22} & \textbf{0.11} & \textbf{0.20} & \textbf{0.31} & \textbf{0.22} & \textbf{0.27} & \textbf{0.81} \\
harddense  & \textbf{0.23} & \textbf{0.85} & \textcolor{gray}{0.26} & \textcolor{gray}{2.77} & \textbf{0.34} & \textbf{0.66} & \textbf{0.28} & \textbf{0.83} & \textbf{0.45} & \textbf{0.66} & \textbf{0.31} & \textbf{0.92} \\
\midrule
\textbf{$\#$ safe}  & \multicolumn{2}{c|}{\textbf{\textcolor{blue}{18}}} & \multicolumn{2}{c|}{\textbf{11}} & \multicolumn{2}{c|}{\textbf{\textcolor{blue}{31}}} & \multicolumn{2}{c|}{\textbf{19}} & \multicolumn{2}{c|}{\textbf{\textcolor{blue}{33}}} & \multicolumn{2}{c}{\textbf{22}} \\
\bottomrule
\end{tabular}
\end{adjustbox}
\end{table*}

\begin{table*}[ht!]
\centering
\caption{Performance across different extra cost configurations for CAPS(SAC+BC). The cost threshold is 1. The $\uparrow$ symbol denotes that a higher reward is better. The $\downarrow$ symbol denotes that a lower cost (up to threshold 1) is better. Each value is averaged over 20 evaluation episodes, and 3 random seeds. {\bf Bold}: Safe agents whose normalized cost $\leq$ 1. \textcolor{gray}{Gray}: Unsafe agents.}
\label{tab:sac_low_costs}
\begin{adjustbox}{max width=\textwidth}
\begin{tabular}{l|cc|cc|cc}
\toprule
\textbf{Cost Configuration} & \multicolumn{2}{c|}{\{5, 10\}} & \multicolumn{2}{c|}{\{15, 30\}} & \multicolumn{2}{c}{\{30, 60\}} \\
\cmidrule(lr){2-3} \cmidrule(lr){4-5} \cmidrule(lr){6-7} 
& reward $\uparrow$ & cost $\downarrow$ & reward $\uparrow$ & cost $\downarrow$ & reward $\uparrow$ & cost $\downarrow$  \\
\midrule
PointCircle1  & \textbf{0.51} & \textbf{0.35} & \textbf{0.52} & \textbf{0.22} & \textbf{0.55} & \textbf{0.51} \\
PointCircle2  & \textcolor{gray}{0.59} & \textcolor{gray}{3.88} & \textcolor{gray}{0.60} & \textcolor{gray}{1.37} & \textcolor{gray}{0.61} & \textcolor{gray}{1.05} \\
SwimmerVelocity  & \textbf{0.32} & \textbf{0.14} & \textbf{0.37} & \textbf{0.47} & \textbf{0.49} & \textbf{0.63} \\
HopperVelocity  & \textcolor{gray}{0.61} & \textcolor{gray}{2.35} & \textbf{0.58} & \textbf{0.75} & \textbf{0.59} & \textbf{0.39} \\
HalfCheetahVelocity  & \textbf{0.88} & \textbf{0.11} & \textbf{0.94} & \textbf{0.19} & \textbf{0.97} & \textbf{0.28} \\
AntVelocity  & \textbf{0.73} & \textbf{0.35} & \textbf{0.86} & \textbf{0.69} & \textbf{0.96} & \textbf{0.73} \\

\midrule
CarCircle  & \textbf{0.56} & \textbf{0.00} & \textbf{0.58} & \textbf{0.02} & \textbf{0.66} & \textbf{0.51} \\
BallRun  & \textcolor{gray}{0.25} & \textcolor{gray}{1.56} & \textbf{0.25} & \textbf{0.67} & \textbf{0.26} & \textbf{0.68} \\
AntRun  & \textcolor{gray}{0.41} & \textcolor{gray}{0.01} & \textbf{0.42} & \textbf{0.01} & \textbf{0.64} & \textbf{0.47} \\

\midrule
EasyDense  & \textbf{0.08} & \textbf{0.39} & \textbf{0.06} & \textbf{0.12} & \textbf{0.09} & \textbf{0.15} \\

MediumMean  & \textbf{0.11} & \textbf{0.66} & \textbf{0.07} & \textbf{0.10} & \textbf{0.10} & \textbf{0.10} \\

hardsparse  & \textbf{0.10} & \textbf{0.38} & \textbf{0.10} & \textbf{0.20} & \textbf{0.11} & \textbf{0.18} \\

\bottomrule
\end{tabular}
\end{adjustbox}
\end{table*}

\subsection{Results for Fitted-Q Evaluation (FQE) Variants of CAPS}\label{ap:fqe}

\vspace{1ex}

The table in Appendix \ref{tab:sfqe_ope} showcases the ablation study on off-policy evaluation (OPE) by providing a detailed comparison of the performance of different Q-value function configurations within the CAPS framework across all environments. The table compares the original CAPS method, which uses offline RL-based $Q^c$ and $Q^r$, with two FQE-based variants: 1) the reward-cost FQE approach, which employs FQE-based $\hat{Q}^r$ and $\hat{Q}^c$ for both reward and cost policies, and 2) the reward FQE method, which uses FQE-based $\hat{Q}^r$ for rewards while retaining $Q^c$ from offline RL for costs.
The results in the table show that the reward-cost FQE method often produces overly conservative policies, leading to lower rewards across various environments. This is because the $\hat{Q}^c$ functions frequently estimate that actions exceed the cost limit, resulting in the selection of low-cost but low-reward actions. In contrast, the reward FQE method, which uses FQE-based $\hat{Q}^r$ combined with the offline RL-based $Q^c$, achieves better performance by maintaining higher rewards while still adhering to safety constraints, as indicated by the blue-highlighted entries in the table.
These findings support the ablation study's conclusion that while FQE can effectively differentiate between actions in terms of rewards, it struggles with the precision required for cost estimation, which is critical for safety in decision-making. As such, the original CAPS method, which relies on offline RL-based $Q^c$ for cost estimation, remains a robust approach for balancing reward maximization and safety. 

\begin{table*}[ht!]
    \centering
    \caption{Performance comparison of FQE based Q-value functions. The cost threshold is 1. The $\uparrow$ symbol denotes that a higher reward is better. The $\downarrow$ symbol denotes that a lower cost (up to threshold 1) is better. Each value is averaged over 3 distinct cost thresholds, 20 evaluation episodes, and 3 random seeds. {\bf Bold:} Safe agents whose normalized cost is smaller than 1. \textcolor{gray}{Gray}: Unsafe agents. \textcolor{blue}{Blue}: Safe agent with the highest reward.}
    \label{tab:sfqe_ope}
    \begin{adjustbox}{max width=\textwidth}
        \begin{tabular}{lcccccc}
        \toprule
         & \multicolumn{2}{c}{FQE $\hat{Q}^r$ $\&$ $\hat{Q}^c$} & \multicolumn{2}{c}{$Q^c$ + FQE $\hat{Q}^r$} & \multicolumn{2}{c}{$Q^c$ $\&$ $Q^r$}  \\
        \cmidrule(lr){2-3} \cmidrule(lr){4-5} \cmidrule(lr){6-7} 
        Task & reward $\uparrow$ & cost $\downarrow$ & reward $\uparrow$ & cost $\downarrow$ & reward $\uparrow$ & cost $\downarrow$  \\
        \midrule
        Bullet Safety Gym & &  &  &  &  &   \\
        \midrule
        AntCircle & \textbf{0.32} & \textbf{0.01} & \textbf{0.39} & \textbf{0.14} & \textbf{\textcolor{blue}{0.41}} & \textbf{\textcolor{blue}{0.15}} \\
        AntRun & \textbf{0.43} & \textbf{0.36} & \textbf{0.6} & \textbf{0.88} & \textbf{\textcolor{blue}{0.61}} & \textbf{\textcolor{blue}{0.90}} \\
        CarCircle & \textbf{0.18} & \textbf{0.14} & \textbf{\textcolor{blue}{0.7}} & \textbf{\textcolor{blue}{0.66}} & \textbf{0.69} & \textbf{0.65} \\
        DroneCircle & \textbf{0.19} & \textbf{0.01} & \textbf{\textcolor{blue}{0.55}} & \textbf{\textcolor{blue}{0.69}} & \textbf{\textcolor{blue}{0.55}} & \textbf{\textcolor{blue}{0.67}} \\
        DroneRun & \textcolor{gray}{0.40} & \textcolor{gray}{3.58} & \textcolor{gray}{0.44} & \textcolor{gray}{2.97} & \textcolor{gray}{0.47} & \textcolor{gray}{2.19} \\
        BallCircle & \textbf{0.41} & \textbf{0.17} & \textbf{\textcolor{blue}{0.69}} & \textbf{\textcolor{blue}{0.59}} & \textbf{\textcolor{blue}{0.69}} & \textbf{\textcolor{blue}{0.59}} \\
        BallRun & \textbf{0.06} & \textbf{0} & \textbf{0.13} & \textbf{0.58} & \textbf{\textcolor{blue}{0.19}} & \textbf{\textcolor{blue}{0.94}} \\
        CarRun & \textbf{0.94} & \textbf{0.16} & \textbf{0.95} & \textbf{0.12} & \textbf{\textcolor{blue}{0.97}} & \textbf{\textcolor{blue}{0.25}} \\

        \midrule
        Safety Gymnasium Car & & & & & &  \\
        \midrule
        CarButton1 & \textbf{\textcolor{blue}{-0.01}} & \textbf{\textcolor{blue}{0.44}} & \textbf{-0.03} & \textbf{0.18} & \textbf{\textcolor{blue}{-0.01}} & \textbf{\textcolor{blue}{0.30}} \\
        CarButton2 & \textbf{\textcolor{blue}{-0.08}} & \textbf{\textcolor{blue}{0.36}} & \textbf{-0.09} & \textbf{0.34} & \textbf{\textcolor{blue}{-0.08}} & \textbf{\textcolor{blue}{0.32}} \\
        CarCircle1 & \textcolor{gray}{0.47} & \textcolor{gray}{1.09} & \textcolor{gray}{0.54} & \textcolor{gray}{1.55} & \textcolor{gray}{0.54} & \textcolor{gray}{1.51} \\
        CarCircle2 & \textcolor{gray}{0.43} & \textcolor{gray}{1.41} & \textcolor{gray}{0.49} & \textcolor{gray}{1.65} & \textcolor{gray}{0.50} & \textcolor{gray}{1.55} \\
        CarGoal1 & \textbf{0.25} & \textbf{0.31} & \textbf{0.29} & \textbf{0.42} & \textbf{\textcolor{blue}{0.33}} & \textbf{\textcolor{blue}{0.38}} \\
        CarGoal2 & \textbf{0.14} & \textbf{0.50} & \textbf{0.15} & \textbf{0.59} & \textbf{\textcolor{blue}{0.16}} & \textbf{\textcolor{blue}{0.62}} \\
        CarPush1 & \textbf{0.17} & \textbf{0.29} & \textbf{\textcolor{blue}{0.20}} & \textbf{\textcolor{blue}{0.37}} & \textbf{\textcolor{blue}{0.20}} & \textbf{\textcolor{blue}{0.31}} \\
        CarPush2 & \textbf{0.03} & \textbf{0.48} & \textbf{\textcolor{blue}{0.09}} & \textbf{\textcolor{blue}{0.67}} & \textbf{0.07} & \textbf{0.75} \\

        \midrule
        Safety Gymnasium Point & & & & & &  \\
        \midrule
        PointButton1 & \textbf{0.01} & \textbf{0.38} & \textbf{0.02} & \textbf{0.40} & \textbf{\textcolor{blue}{0.03}} & \textbf{\textcolor{blue}{0.45}} \\
        PointButton2 & \textbf{0.09} & \textbf{0.62} & \textbf{0.13} & \textbf{0.94} & \textbf{\textcolor{blue}{0.14}} & \textbf{\textcolor{blue}{0.75}} \\
        PointCircle1 & \textbf{0.23} & \textbf{0.37} & \textcolor{gray}{0.44} & \textcolor{gray}{1.22} & \textbf{\textcolor{blue}{0.50}} & \textbf{\textcolor{blue}{0.78}} \\
        PointCircle2 & \textbf{0.40} & \textbf{0.21} & \textbf{0.50} & \textbf{0.77} & \textbf{\textcolor{blue}{0.51}} & \textbf{\textcolor{blue}{0.80}} \\
        PointGoal1 & \textbf{0.25} & \textbf{0.36} & \textbf{0.37} & \textbf{0.42} & \textbf{\textcolor{blue}{0.47}} & \textbf{\textcolor{blue}{0.53}} \\
        PointGoal2 & \textbf{0.14} & \textbf{0.38} & \textbf{0.25} & \textbf{0.53} & \textbf{\textcolor{blue}{0.33}} & \textbf{\textcolor{blue}{0.80}} \\
        PointPush1 & \textbf{0.14} & \textbf{0.21} & \textbf{0.16} & \textbf{0.38} & \textbf{\textcolor{blue}{0.19}} & \textbf{\textcolor{blue}{0.29}} \\
        PointPush2 & \textbf{0.11} & \textbf{0.56} & \textbf{0.12} & \textbf{0.60} & \textbf{\textcolor{blue}{0.13}} & \textbf{\textcolor{blue}{0.64}} \\
        
        \midrule
        Safety Gymnasium Velocity & & & & & & \\
        \midrule
        AntVelocity & \textbf{0.87} & \textbf{0.18} & \textbf{\textcolor{blue}{0.95}} & \textbf{\textcolor{blue}{0.7}} & \textbf{\textcolor{blue}{0.95}} & \textbf{\textcolor{blue}{0.64}} \\
        HalfCheetahVelocity & \textbf{0.87} & \textbf{0.18} & \textbf{\textcolor{blue}{0.94}} & \textbf{\textcolor{blue}{0.66}} & \textbf{\textcolor{blue}{0.94}} & \textbf{\textcolor{blue}{0.77}} \\
        HopperVelocity & \textbf{0.24} & \textbf{0.25} & \textbf{\textcolor{blue}{0.43}} & \textbf{\textcolor{blue}{0.72}} & \textbf{0.41} & \textbf{0.70} \\
        SwimmerVelocity & \textbf{\textcolor{blue}{0.4}} & \textbf{\textcolor{blue}{0.68}} & \textcolor{gray}{0.43} & \textcolor{gray}{2.02} & \textcolor{gray}{0.43} & \textcolor{gray}{1.58} \\
        Walker2dVelocity & \textbf{\textcolor{blue}{0.76}} & \textbf{\textcolor{blue}{0.06}} & \textbf{0.71} & \textbf{0.72} & \textbf{0.80} & \textbf{0.62} \\
        
        \midrule
        Metadrive environments & & & & & &  \\
        \midrule
        easysparse & \textcolor{gray}{0.24} & \textcolor{gray}{1.39} & \textbf{\textcolor{blue}{0.11}} & \textbf{\textcolor{blue}{0.33}} & \textbf{\textcolor{blue}{0.11}} & \textbf{\textcolor{blue}{0.34}} \\
        easymean & \textbf{0} & \textbf{0.16} & \textbf{\textcolor{blue}{0.01}} & \textbf{\textcolor{blue}{0.20}} & \textbf{\textcolor{blue}{0.01}} & \textbf{\textcolor{blue}{0.20}} \\
        easydense & \textbf{\textcolor{blue}{0.12}} & \textbf{\textcolor{blue}{0.11}} & \textbf{\textcolor{blue}{0.12}} & \textbf{\textcolor{blue}{0.18}} & \textbf{\textcolor{blue}{0.10}} & \textbf{\textcolor{blue}{0.19}} \\
        mediumsparse & \textbf{0.43} & \textbf{0.34} & \textbf{0.56} & \textbf{0.73} & \textbf{\textcolor{blue}{0.60}} & \textbf{\textcolor{blue}{0.74}} \\
        mediummean & \textbf{0.55} & \textbf{0.83} & \textbf{\textcolor{blue}{0.66}} & \textbf{\textcolor{blue}{0.86}} & \textbf{\textcolor{blue}{0.66}} & \textbf{\textcolor{blue}{0.94}} \\
        mediumdense & \textbf{\textcolor{blue}{0.76}} & \textbf{\textcolor{blue}{0.60}} & \textbf{0.67} & \textbf{0.61} & \textbf{0.69} & \textbf{0.56} \\
        hardsparse & \textbf{0.42} & \textbf{0.51} & \textbf{\textcolor{blue}{0.45}} & \textbf{\textcolor{blue}{0.76}} & \textbf{\textcolor{blue}{0.45}} & \textbf{\textcolor{blue}{0.72}} \\
        hardmean & \textbf{0.20} & \textbf{0.06} & \textbf{0.23} & \textbf{0.06} & \textbf{\textcolor{blue}{0.28}} & \textbf{\textcolor{blue}{0.25}} \\
        harddense & \textbf{0.28} & \textbf{0.39} & \textbf{0.30} & \textbf{0.53} & \textbf{\textcolor{blue}{0.37}} & \textbf{\textcolor{blue}{0.67}} \\
        
        \bottomrule
    \end{tabular}

\end{adjustbox}
\end{table*}

\section{Experimental Details}

\vspace{2ex}

\subsection{Environment descriptions} \label{ap:envs_desc}
The environments designed for testing safe offline reinforcement learning (RL) methods are built on different simulators, each tailored to specific tasks and agents.

\vspace{1ex}

\begin{figure*}[ht]
    \centering
    \includegraphics[scale=0.3]{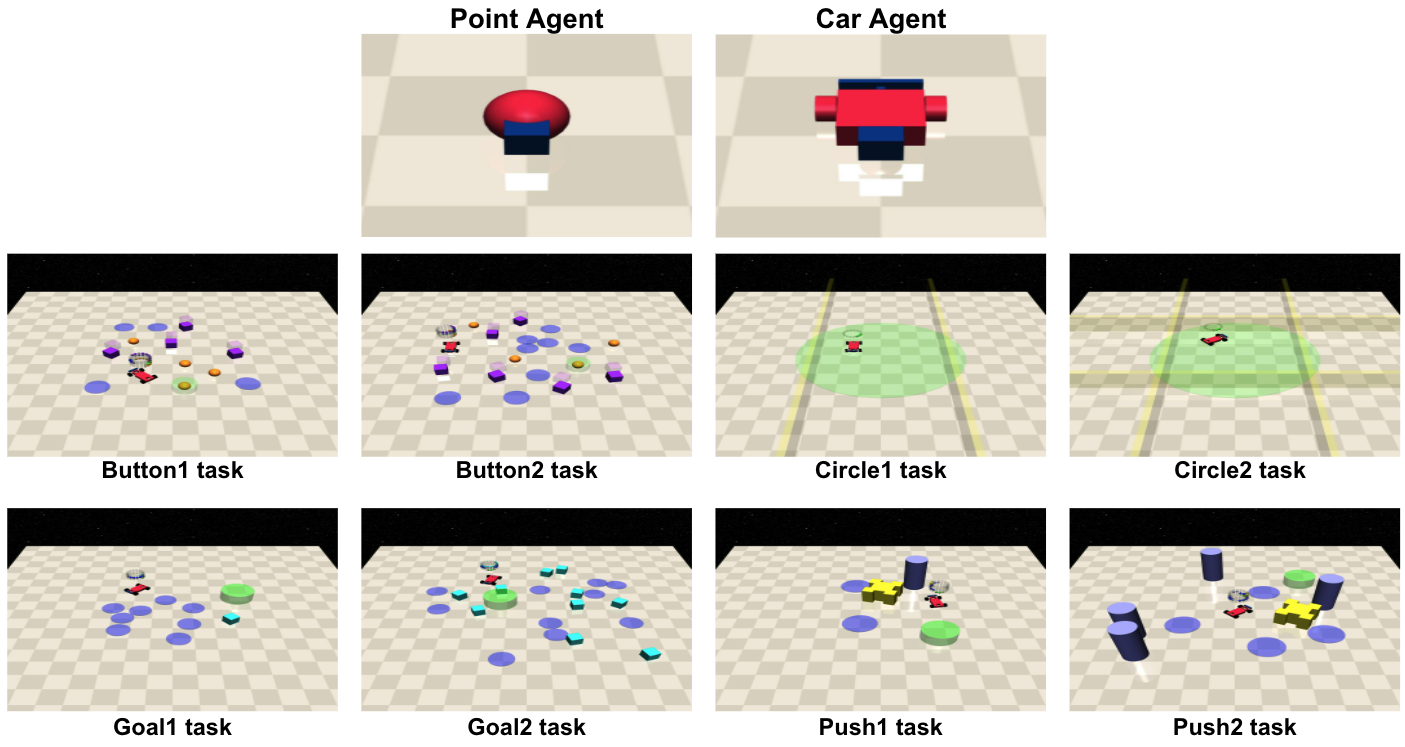}
    \caption{Visualization of the Safety-Gymnasium environments.}
    \label{fig:safety_gym}
\end{figure*}

\noindent \textbf{Safety-Gymnasium \cite{ray2019benchmarking, ji2023omnisafe}:} Developed using the Mujoco physics simulator, Safety-Gymnasium offers a variety of environments focused on safety-critical tasks. The Car agent has several tasks, including Button, Push, and Goal, each available in two difficulty levels, requiring the agent to navigate hazards while achieving specific objectives. For instance, in the Goal task, the agent moves towards multiple goal positions, with each new goal randomly reset upon completion. In the Push task, the agent must move a box to different goal positions, with new locations generated after each success. The Button task involves navigating to and pressing goal buttons scattered across the environment. Additionally, Safety-Gymnasium includes velocity constraint tasks for agents such as Ant, HalfCheetah, and Swimmer. In the Velocity task, the robot coordinates leg movement to move forward, while in the Run task, it starts in a random direction and speed to reach the opposite side of the map. The Circle task rewards agents for following a path along a green circle while avoiding the red region outside. Tasks are named by combining the agent, task, and difficulty level (e.g., CarPush1), reflecting the complexity and objectives of each scenario. Figure \ref{fig:safety_gym} illustrates these tasks.

\vspace{1ex}

\begin{figure*}[ht]
    \centering
    \includegraphics[scale=0.3]{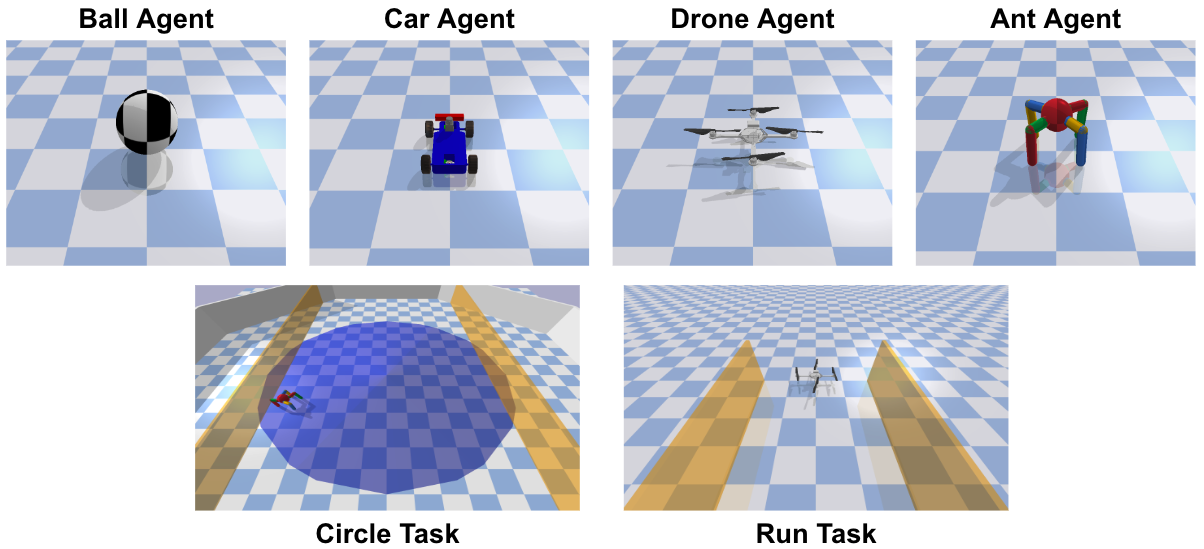}
    \caption{Visualization of the Bullet-Safety-Gym environments.}
    \label{fig:py_bullet}
\end{figure*}

\noindent \textbf{Bullet-Safety-Gym \cite{gronauer2022bullet}:} This environment suite, created using the PyBullet physics simulator, includes four agent types—Ball, Car, Drone, and Ant—along with two primary tasks: Circle and Run. In the Run task, agents navigate through a corridor bounded by safety lines that they can cross without physical collision, though doing so incurs a penalty. Additionally, if an agent exceeds a certain speed limit, it accrues extra penalties. In the Circle task, agents are required to move clockwise around a circular path, with rewards increasing as they maintain higher speeds closer to the boundary. Penalties are given if the agent strays outside a predefined safety zone. These environments are designed to evaluate offline reinforcement learning methods with a focus on safety, featuring shorter, more straightforward tasks when compared to those in Safety-Gymnasium. Figure \ref{fig:py_bullet} provides a visual representation of these tasks.

\vspace{1ex}

\begin{figure*}[ht]
    \centering
    \includegraphics[scale=0.3]{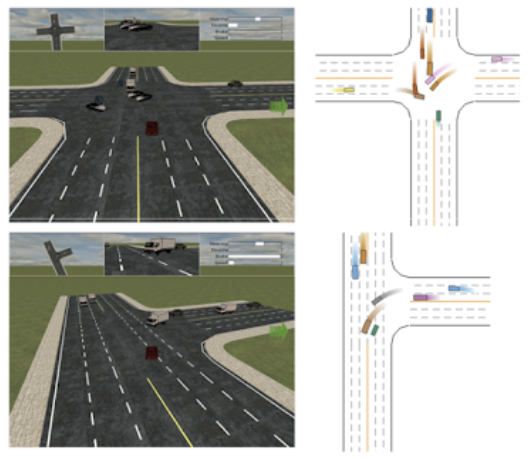}
    \caption{Visualization of the MetaDrive environments.}
    \label{fig:metadrive}
\end{figure*}

\noindent \textbf{MetaDrive \cite{li2022metadrive}:} MetaDrive is a self-driving simulation environment based on the Panda3D game engine. It replicates realistic driving conditions with varying levels of road complexity (easy, medium, hard) and traffic density (sparse, mean, dense). Tasks are named according to the combination of road and vehicle conditions. This environment allows the assessment of offline RL algorithms in scenarios that mirror real-world driving challenges. Figure \ref{fig:metadrive} visualizes these tasks.

Each of these environments offers unique challenges for testing offline safe RL methods, from driving simulations to tasks that require careful navigation and hazard avoidance. 

\subsection{Computation Time} \label{ap:computation_time}
The computation time comparison in Table \ref{tab:computation_time} reveals that both IQL and SAC+BC instantiations of CAPS training are significantly more efficient than CDT training, tested on the HalfCheetah task. While CDT takes approximately 154 minutes, CAPS(IQL) completes the task in a much shorter time, ranging from 24 to 33 minutes, depending on the number of heads. Similarly, CAPS(SAC+BC), though slightly more time-intensive than IQL, still remains considerably faster than CDT, with computation times ranging from 44 to 103 minutes. This demonstrates the training efficiency of both IQL and SAC+BC instantiations, particularly when compared to CDT. All experiments were performed on an A40 GPU and an AMD EPYC 7573X 32-Core Processor.
\begin{table*}[ht]
    \centering
    \caption{Approximate computation time comparison on Halfcheetah task (NVIDIA A40)}
    \begin{tabular}{lc}
        \toprule
        \textbf{Method} & \textbf{Time (min)} \\
        \midrule
        \textbf{CDT} & $\approx$ 154 \\
        \midrule
        \textbf{CAPS IQL} & \\
        \quad 2 heads & $\approx$ 24 \\
        \quad 4 heads & $\approx$ 26 \\
        \quad 8 heads & $\approx$ 33 \\
        \midrule
        \textbf{CAPS SAC+BC}  & \\
        \quad 2 heads & $\approx$ 44 \\
        \quad 4 heads & $\approx$ 62 \\
        \quad 8 heads & $\approx$ 103\\
        \bottomrule
    \end{tabular}
    \label{tab:computation_time}
\end{table*}

\subsection{CAPS Hyperparameters} \label{ap:caps_hyper}
\vspace{1ex}

Table \ref{tab:caps_hyper} presents the hyperparameters used across different environments (BulletGym, SafetyGym, and MetaDrive) for  CAPS, detailing common configurations such as cost thresholds, training steps, and network specifications.
\begin{table*}[ht]
\centering
\caption{CAPS hyperparameters.}
\begin{tabular}{lccc}
\midrule
\textbf{Common Parameters} & \textbf{BulletGym} & \textbf{SafetyGym} & \textbf{MetaDrive} \\ 
\midrule
Cost thresholds (for evaluation) & \{10, 20, 40\} & \{20, 40, 80\} & \{10, 20, 40\} \\ 
Training steps & \multicolumn{2}{c}{100000} & 200000 \\ 
discount $\gamma$ & \multicolumn{3}{c}{0.99} \\
Batch size & \multicolumn{3}{c}{512} \\ 
Optimizer & \multicolumn{3}{c}{Adam} \\ 
Actor, Critic, and Value Networks hidden size & \multicolumn{3}{c}{\{512, 512\}}  \\ 
$\lambda_k$ & \multicolumn{3}{c}{$k \big/ ((\text{heads\_count} - 1)/2) $} \\
seeds & \multicolumn{3}{c}{\{0, 10, 20\}} \\
\midrule
\textbf{IQL instantiation} & & & \\
\midrule
IQL Expectile $\tau$ \{$V_r$, $V_c$\}  & \multicolumn{2}{c}{\{0.7, 0.7\}}  & \{0.5, 0.5\} \\
IQL Inverse Temperature $\beta$   & \multicolumn{3}{c}{3} \\ 
Actor, Critic, and Value Networks learning rate & \multicolumn{3}{c}{3e-4} \\ 
 
\midrule
\textbf{SAC+BC instantiation} & & & \\
\midrule
Actor and learning rate & \multicolumn{3}{c}{1e-4} \\
Critic learning rate & \multicolumn{3}{c}{1e-3} \\
\end{tabular}
\label{tab:caps_hyper}
\end{table*}

\end{document}